\documentclass[11pt,a4paper,gdm]{google}

\usepackage[authoryear,sort&compress,round]{natbib}
\usepackage[nopatch=footnote]{microtype}
\usepackage{amsmath,amsfonts,amssymb}
\usepackage{graphicx}
\usepackage[dvipsnames]{xcolor}
\usepackage{colortbl}
\usepackage{booktabs}
\usepackage{tabularx}
\usepackage{nicefrac}
\usepackage{multirow}
\usepackage{tikz}
\usepackage{pgfplots}
\usepackage{float}
\usepackage{wrapfig}
\usepackage{subcaption}
\usepackage[normalem]{ulem}
\usepackage{needspace}
\usepackage[colorlinks=true,allcolors=blue]{hyperref}
\usepackage{url}
\hypersetup{
  pdftitle={Scaffolding Minds: Optimizing Latent Visual Target Representations for Multimodal Reasoning},
  pdfauthor={Haoqiang Kang, Yinpeng Chen, Luyang Liu, Jesper Sparre Andersen, Abhijit Ogale, Baochen Sun, Lichan Hong, Ed H. Chi}
}
\pgfplotsset{compat=1.18}
\usetikzlibrary{positioning, shadows, arrows.meta, decorations.pathreplacing}
\usepgfplotslibrary{fillbetween}

\definecolor{sftgray}{RGB}{149,165,166}
\definecolor{bestblue}{RGB}{52,152,219}
\definecolor{oursred}{RGB}{211,47,47}
\definecolor{diffgreen}{RGB}{39,174,96}

\definecolor{headerbg}{RGB}{228,236,248}
\definecolor{basebg}{RGB}{248,248,248}
\definecolor{oursbg}{RGB}{253,236,236}
\definecolor{oursbest}{RGB}{250,220,220}
\definecolor{deltabg}{RGB}{232,246,232}
\definecolor{gaingreen}{RGB}{27,130,60}

\newif\ifshowrev \showrevfalse
\ifshowrev
  
  \newcommand{\old}[1]{{\color{blue}\sout{#1}}}

\else
  
  \newcommand{\old}[1]{}

\fi

\ifshowrev
  \newcommand{\rtag}{\textsuperscript{\textbf{[R2]}}\,}
  
  \newcommand{\rtwodel}[1]{{\color{blue}\rtag\sout{#1}}}
  
  \newcommand{\rtwonote}[1]{{\color{blue}\small\textbf{[R2 removed:} #1\textbf{]}}}
  
\else
  \newcommand{\rtag}{}
  
  \newcommand{\rtwodel}[1]{}
  
  \newcommand{\rtwonote}[1]{}
  
\fi
\ifshowrev

  \newcommand{\rnd}[1]{{\color{blue}\textsuperscript{\textbf{[R19]}}\,\sout{#1}}}
\else

  \newcommand{\rnd}[1]{}
\fi
\ifshowrev
  
  \newcommand{\reighteendel}[1]{{\color{blue}\textsuperscript{\textbf{[R18]}}\,\sout{#1}}}
\else
  
  \newcommand{\reighteendel}[1]{}
\fi
\ifshowrev
  
  \newcommand{\rseventeendel}[1]{{\color{blue}\textsuperscript{\textbf{[R17]}}\,\sout{#1}}}
\else
  
  \newcommand{\rseventeendel}[1]{}
\fi
\ifshowrev

\else

\fi
\ifshowrev
  
  \newcommand{\rsixteendel}[1]{{\color{blue}\textsuperscript{\textbf{[R16]}}\,\sout{#1}}}
\else
  
  \newcommand{\rsixteendel}[1]{}
\fi
\ifshowrev
  
  \newcommand{\rthirteendel}[1]{{\color{blue}\textsuperscript{\textbf{[R13]}}\,\sout{#1}}}
\else
  
  \newcommand{\rthirteendel}[1]{}
\fi
\ifshowrev
  
  \newcommand{\rtwelvenote}[1]{{\color{blue}\small\textbf{[R12:} #1\textbf{]}}}
\else
  
  \newcommand{\rtwelvenote}[1]{}
\fi
\ifshowrev
  
  \newcommand{\relevennote}[1]{{\color{blue}\small\textbf{[R11:} #1\textbf{]}}}
\else
  
  \newcommand{\relevennote}[1]{}
\fi
\ifshowrev
  
  \newcommand{\rtendel}[1]{{\color{blue}\textsuperscript{\textbf{[R10]}}\,\sout{#1}}}
\else
  
  \newcommand{\rtendel}[1]{}
\fi
\ifshowrev
  
\else
  
\fi
\ifshowrev
  
  \newcommand{\reightdel}[1]{{\color{blue}\textsuperscript{\textbf{[R8]}}\,\sout{#1}}}
  \newcommand{\reightnote}[1]{{\color{blue}\small\textbf{[R8:} #1\textbf{]}}}
\else
  
  \newcommand{\reightdel}[1]{}
  \newcommand{\reightnote}[1]{}
\fi
\ifshowrev
  
  \newcommand{\rsevendel}[1]{{\color{blue}\textsuperscript{\textbf{[R7]}}\,\sout{#1}}}
\else
  
  \newcommand{\rsevendel}[1]{}
\fi
\ifshowrev
  \newcommand{\rsixmove}[1]{\marginpar{\raggedright\color{red}\scriptsize\textbf{[R6]} #1}}
\else
  \newcommand{\rsixmove}[1]{}
\fi
\ifshowrev
  
  \newcommand{\rfivemove}[1]{\marginpar{\raggedright\color{red}\scriptsize\textbf{[R5]} #1}}
  
\else
  
  \newcommand{\rfivemove}[1]{}
  
\fi
\ifshowrev
  
  \newcommand{\rfournote}[1]{{\color{blue}\small\textbf{[R4:} #1\textbf{]}}}
\else
  
  \newcommand{\rfournote}[1]{}
\fi

\ifshowrev
  \newcommand{\rthreetag}{\textsuperscript{\textbf{[R3]}}\,}
  
  \newcommand{\rthreedel}[1]{{\color{blue}\rthreetag\sout{#1}}}
  \newcommand{\rthreenote}[1]{{\color{blue}\small\textbf{[R3:} #1\textbf{]}}}
\else
  \newcommand{\rthreetag}{}
  
  \newcommand{\rthreedel}[1]{}
  \newcommand{\rthreenote}[1]{}
\fi

\title{Scaffolding Minds: Optimizing Latent Visual Target Representations for Multimodal Reasoning}

\uselogo{}

\author[1,2]{Haoqiang Kang}
\author[1]{Yinpeng Chen}
\author[1,*]{Luyang Liu}
\author[1]{Jesper Sparre Andersen}
\author[1]{Abhijit Ogale}
\author[1]{Baochen Sun}
\author[1]{Lichan Hong}
\author[1]{Ed H. Chi}
\affil[1]{Google DeepMind}
\affil[2]{UC San Diego}
\affil[*]{Work done at Google DeepMind.}

\newcommand{\answerYes}[1][]{[Yes]#1}
\newcommand{\answerNo}[1][]{\textcolor{orange}{[No]#1}}
\newcommand{\answerNA}[1][]{\textcolor{gray}{[N/A]#1}}

\begin{abstract}
  Latent reasoning has advanced multimodal reasoning through a two-stage training paradigm: (1) a helper image is encoded into latent tokens to teach visual chain-of-thought during a \emph{supervised fine-tuning (SFT) stage}, and (2) these latent tokens are further refined with reward feedback during a \emph{reinforcement learning (RL) stage}. In this paper, we identify two key limitations of this framework, one in each stage. First, the SFT stage typically relies on an off-the-shelf vision encoder to encode the helper image, yielding suboptimal latent representations that may not be well aligned with the downstream reasoning task.
  Second, existing RL methods treat the latent component only through deterministic regularization, which constrains policy drift but does not create alternative latent trajectories for exploration.
  To address these limitations, we propose \textbf{Scaffolding Minds}.
  Our approach learns a dedicated scaffolding encoder that provides an optimized target in latent space, and learns both the mean and variance of the RL sampler.
  We further show that these two improvements are complementary, together yielding substantial gains over strong baselines. Empirically, our method improves over the strongest latent-reasoning baseline by \textbf{+9.5} points on FrozenLake spatial planning, with the gain widening to \textbf{+19} points on the $32{\times}32$ grids, and by \textbf{+5.6} points on average across nine visual-centric reasoning benchmarks.
\end{abstract}

\begin{document}
\raggedbottom

\maketitle

\begin{figure}[!h]
  \centering
  \includegraphics[width=0.95\linewidth]{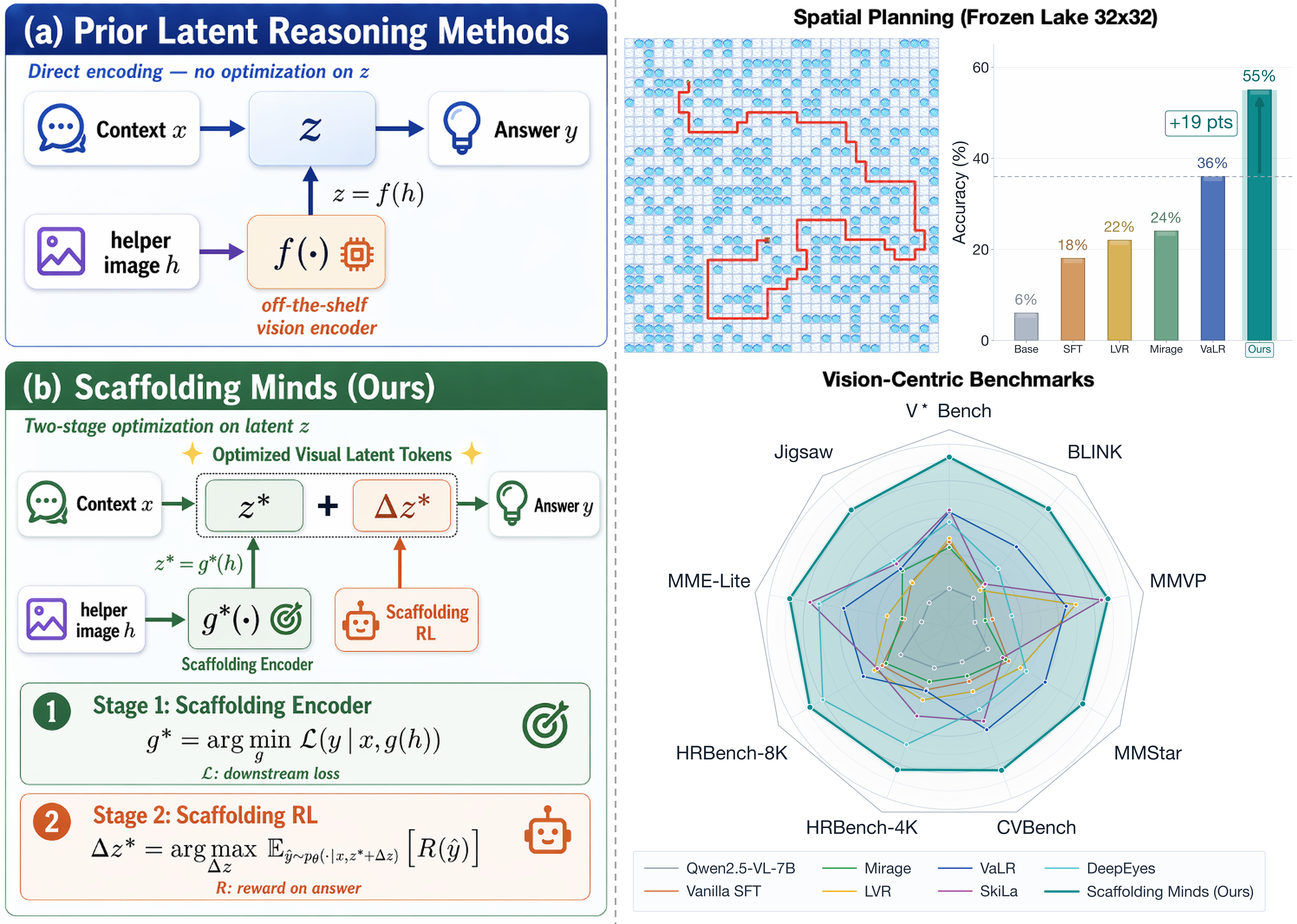}
  \caption{\textbf{Scaffolding Minds.} \textbf{Left:} Prior latent-reasoning methods derive latent targets by encoding a training-time helper image with a frozen off-the-shelf vision encoder. Our method instead learns a dedicated scaffolding encoder that optimizes the latent target for downstream reasoning, then uses a learned Gaussian policy to sample reward-guided residual latent actions for direct exploration. \textbf{Right:} Scaffolding Minds improves the strongest prior latent baseline by $+9.5$ points on average across FrozenLake levels, with the gain widening to $+19$ points on the hardest $32{\times}32$ grids, and by $+5.6$ points on average across nine visual-centric reasoning benchmarks.}
  \label{fig:teaser}
\end{figure}

\section{Introduction}
\label{sec:intro}

Vision-language models (VLMs) have become increasingly capable at multimodal reasoning~\citep{qwen2025vl, liu2024llava, team2023gemini}. Methods such as \citet{wei2022, kojima2022zeroshot, xu2024, zhao2024} rely on textual chain-of-thought, but lose important visual details~\citep{wu2024vstar, fu2024blink, tong2024mmvp}. To bridge this gap, latent visual reasoning~ \citep{yang2025b, li2025lvr, wang2025monet, jeon2026valr, qin2025covt, zhang2025sketchpad, tong2025skila, chen2025} has emerged as a compelling solution. By interleaving small blocks of latent visual tokens directly into the reasoning chain, these models can ``think'' in both text and vision. Typically, these methods follow a two-stage training paradigm. 
In Stage 1 (SFT), they leverage a \emph{helper image} (e.g., an annotated crop) that is available only during training to provide guidance in the latent space (encoded by an off-the-shelf vision encoder). In Stage 2, they use reinforcement learning (RL) to refine both latent and text generation.

\begin{wrapfigure}{r}{0.45\textwidth}
\vspace{-12mm}
  \centering
  \captionsetup{font=small}
  \includegraphics[width=\linewidth]{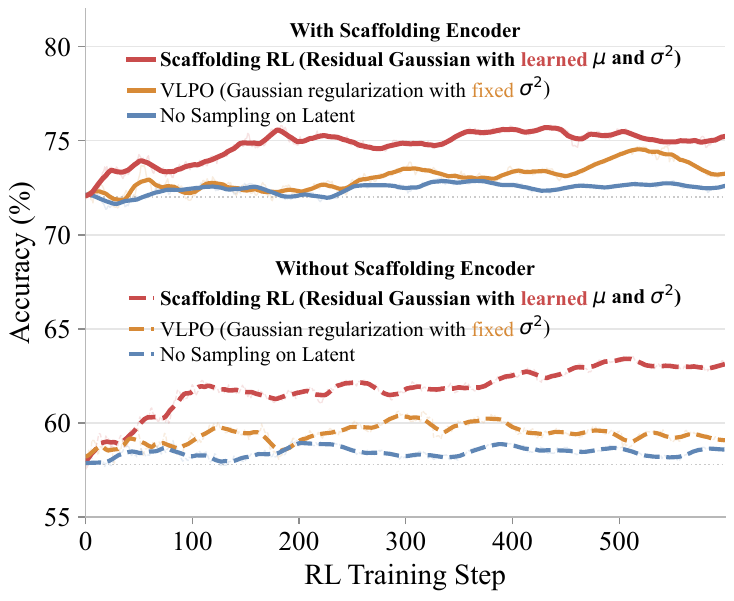}
  \caption{RL performance comparison on the FrozenLake benchmark.}
  \label{fig:rl_full}
  \vspace{-10mm}
\end{wrapfigure}

We identify a limitation at each stage of this paradigm. First, in the SFT stage, the latent target is obtained by encoding the helper image with an off-the-shelf vision encoder. Because this encoder is trained for general-purpose visual representation rather than the downstream reasoning objective, its features may preserve perceptual content that is irrelevant to the task while failing to emphasize the intermediate evidence needed for reasoning. The latent generator is therefore supervised toward a convenient but suboptimal target, limiting how much useful reasoning information can be learned even when the downstream model is trainable. Second, existing RL methods either update only text tokens or apply deterministic regularization to the latent block. Text-only objectives provide no direct likelihood or credit assignment for latent actions, whereas deterministic regularization only constrains how far a latent representation drifts from a reference. Without sampling alternative latent actions for exploration, reward optimization cannot discover more useful latent reasoning trajectories, yielding small or unstable gains as shown in Figure~\ref{fig:rl_full}.

In this paper, we present \textbf{Scaffolding Minds} (Figure~\ref{fig:teaser}, left) to address these two issues. First, for the SFT stage, instead of using fixed off-the-shelf features as the target, we \textit{learn} a \emph{scaffolding encoder} to encode a helper image into a latent target optimized for the reasoning task.

Second, we introduce \emph{Scaffolding RL}, which replaces deterministic latent regularization with an explicit Gaussian sampler over residual latent actions. Two lightweight heads learn its input-adaptive mean and variance, and the policy samples these actions during rollout to explore alternative latent reasoning trajectories (Figure~\ref{fig:rl_full}).
Moreover, we find the scaffolding encoder and scaffolding RL are complementary, yielding performance boosts when combined.

We evaluate \textit{Scaffolding Minds} on FrozenLake spatial planning (across grid sizes from $8{\times}8$ to $32{\times}32$) and nine visual-centric reasoning benchmarks: V$^\star$~\citep{wu2024vstar}, BLINK~\citep{fu2024blink}, MMVP~\citep{tong2024mmvp}, MMStar~\citep{chen2024mmstar}, CVBench~\citep{tong2024cvbench}, HRBench-4K and HRBench-8K~\citep{wang2024hrbench}, MME-RealWorld-Lite~\citep{zhang2025mmerealworld}, and Jigsaw~\citep{lyu2025jigsaw}. Across all suites, our method consistently outperforms the strongest prior latent baseline. On FrozenLake, it raises average accuracy by $+9.5$ points, with a gain of $+19$ points on the most challenging $32{\times}32$ grids. Across the nine visual-centric reasoning benchmarks, we achieve an average accuracy increase of $+5.6$ points.

\section{Preliminary: Latent Reasoning for VLMs}
\label{sec:prelim}

\emph{Latent visual reasoning}~ \citep{yang2025b, li2025lvr, wang2025monet, jeon2026valr} inserts a block of dedicated latent tokens $\mathbf{z}$ between $\langle$BOT$\rangle$ and $\langle$EOT$\rangle$ inside the VLM's chain-of-thought, with their target set to the vision embedding of a helper image $\mathbf{h}$ produced by an off-the-shelf vision encoder $f$. The helper image $\mathbf{h}$ is only available at training time. The reasoning sequence takes the form
$[\mathbf{x},\allowbreak\,\mathbf{c}^{(1)},\allowbreak\,\langle\mathrm{BOT}\rangle,\allowbreak\,\mathbf{z}^{(1)},\allowbreak\,\langle\mathrm{EOT}\rangle,\allowbreak\,\ldots,\allowbreak\,\mathbf{c}^{(B)},\allowbreak\,\langle\mathrm{BOT}\rangle,\allowbreak\,\mathbf{z}^{(B)},\allowbreak\,\langle\mathrm{EOT}\rangle,\allowbreak\,\mathbf{c}^{(B+1)},\allowbreak\,\mathbf{y}]$,
where $\mathbf{x}$ is the input image and query, each $\mathbf{c}^{(b)}$ is optional text reasoning, each $\mathbf{z}^{(b)}\in\mathbb{R}^{K\times d}$ holds $K$ latent tokens, $B\geq1$ is the number of latent blocks, and $\mathbf{y}$ is the answer. We write $\mathbf{z}$ when referring to a single block. As in prior work~\citep{wang2025monet}, the VLM itself decides when to enter latent reasoning by emitting $\langle$BOT$\rangle$, and it returns to text after the block, so a reasoning chain may contain a single latent block or several blocks interleaved with text. Existing pipelines train the VLM in two stages: an \emph{SFT stage} and an \emph{RL stage}.

\paragraph{SFT stage.} The goal of the SFT stage is to teach the VLM to generate the latent block $\mathbf{z}$ from the input image and text query alone (without the helper image), so that no helper image is needed at inference. Essentially, a frozen off-the-shelf vision encoder $f$ maps $\mathbf{h}$ to a target latent block, and the VLM is trained to match its generated latent tokens to this target while jointly fitting the answer $\mathbf{y}$ under next-token prediction.

{\paragraph{RL stage.} The SFT-stage model is further refined under a task reward (Figure~\ref{fig:rl_full}). Existing latent-reasoning RL methods handle the text and latent components differently.

\emph{Text-centric policy optimization.} Standard recipes apply GRPO~\citep{shao2024grpo} to the language-modeling head, restricting policy optimization to the discrete text trajectory~\citep{li2025lvr}. Given input $\mathbf{x}$, the VLM samples $G$ text trajectories and optimizes
\[
\textstyle \mathcal{L}_{\mathrm{text\mbox{-}RL}}(\theta)=-\frac{1}{G}\sum_{i=1}^{G}\frac{1}{T_i}\sum_{t=1}^{T_i}\min\!\big(\rho_{i,t}(\theta)A_i,\,\operatorname{clip}(\rho_{i,t}(\theta),1{-}\epsilon,1{+}\epsilon)A_i\big),
\]

where $\rho_{i,t}(\theta)$ is the current-to-old probability ratio for text token $t$, $A_i$ is the group-relative advantage, and $\epsilon$ is the clipping threshold. Because this objective defines policy probabilities only over discrete text tokens, it neither samples nor directly explores alternative actions in the continuous latent block.

\emph{Visual-Latent Policy Optimization (VLPO)}~\citep{wang2025monet}. VLPO additionally incorporates deterministic latent embeddings through a fixed-variance Gaussian formulation. For an old-policy rollout latent $\mathbf{z}^{\mathrm{old}}_{i,t}$ and the current-policy deterministic output $\mathbf{z}^{\theta}_{i,t}$, its latent factor is
\[
\textstyle r_{i,t}(\theta)=\exp\!\big(-\|\mathbf{z}^{\mathrm{old}}_{i,t}-\mathbf{z}^{\theta}_{i,t}\|_2^2/(2\sigma_0^2)\big).
\]

Here, the current-policy output is the Gaussian mean, the old-policy rollout latent is the point being evaluated, and the variance $\sigma_0^2$ is fixed. Thus, the Gaussian acts as latent regularization that penalizes drift between deterministic latent outputs. It is not a rollout distribution: VLPO does not sample a latent action from this Gaussian and thus creates no alternative latent trajectories for exploration.
}



\section{Method: Scaffolding Minds}
\label{sec:method}
This section first identifies two limitations of existing latent visual reasoning pipelines, then presents \textbf{Scaffolding Minds}, our two-stage framework that addresses both.
\vspace{0mm}
\subsection{Motivation}
\vspace{0mm}

{Existing latent visual reasoning pipelines suffer from two limitations: (1) at the SFT stage, the latent target is produced by a \emph{frozen} off-the-shelf vision encoder $f$, which is not optimized to capture reasoning-sensitive features from the helper image $\mathbf{h}$; and (2) at the RL stage, existing methods either leave the latent block outside RL or use a deterministic, fixed-variance Gaussian formulation, without sampling latent actions for exploration.

Both limitations share a common root cause: the latent target is not optimized for the reasoning task, and RL does not directly explore alternative latent trajectories. To fill this gap, we propose \textbf{Scaffolding Minds} (Figure~\ref{fig:architecture}): a \emph{scaffolding encoder} that learns the latent target end-to-end through the downstream task loss, and \emph{Scaffolding RL} that samples residual latent actions from an input-adaptive Gaussian whose mean and variance are conditioned on the generated latent.
}

\subsection{Stage 1: SFT via Scaffolding Encoder}
\begin{figure}[!t]
  \centering
  \includegraphics[width=\linewidth]{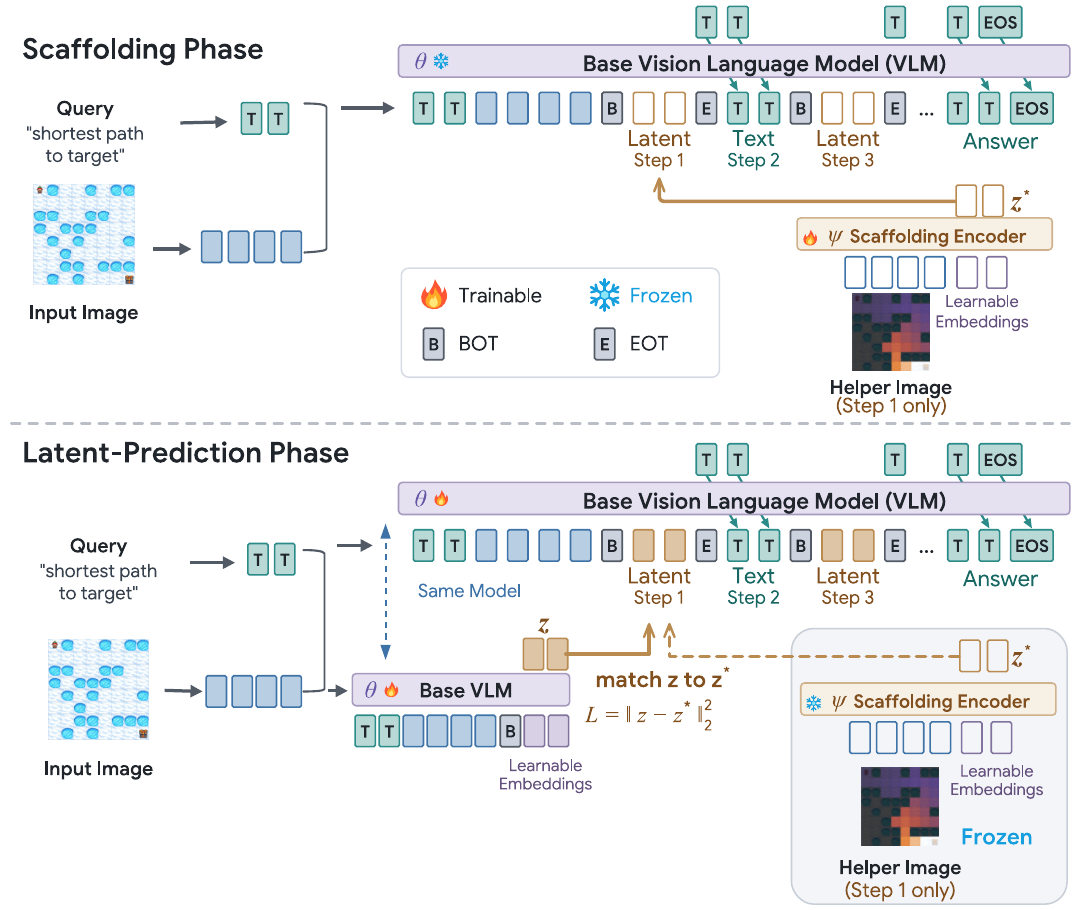}
  \caption{\textbf{SFT stage of Scaffolding Minds.} \textbf{Scaffolding Phase}~(top): the scaffolding encoder $\psi$ is trained to compress the helper image into target latent tokens $\mathbf{z}^{*}$ that drive the frozen base VLM to produce the correct answer. \textbf{Latent-Prediction Phase}~(bottom): the VLM ($\theta$) is then trained to predict latent tokens $\mathbf{z}$ from the input $\mathbf{x}$ alone, supervised by $\ell_2$ matching $\mathcal{L} = \|\mathbf{z} - \mathbf{z}^{*}\|_2^2$ to the (now frozen) scaffolding encoder's targets, while the helper image is no longer used at inference.}
  \label{fig:architecture}
\end{figure}

The SFT stage is to learn an optimized latent target representation $\mathbf{z}^* \in \mathbb{R}^{K \times d}$ from the helper image $\mathbf{h}$, and then use $\mathbf{z}^*$ as supervision to teach the VLM to produce its own latent block $\mathbf{z}$ from the input $\mathbf{x}$ alone, so that no helper image is needed at inference. As shown in Figure~\ref{fig:architecture}, we split this into two phases: (i) the \textit{Scaffolding Phase} trains a dedicated scaffolding encoder with model parameters $\psi$ to map $\mathbf{h}$ to the target $\mathbf{z}^*$, optimized end-to-end through the frozen base VLM's task loss; and (ii) the \textit{Latent-Prediction Phase} fine-tunes the VLM to regress its own latent block $\mathbf{z}$ toward the (now frozen) target $\mathbf{z}^*$, while jointly fitting the answer $\mathbf{y}$ under next-token prediction. After Stage~1, the scaffolding encoder $\psi$ and helper image $\mathbf{h}$ are both discarded; only the VLM is used at inference.

\paragraph{Scaffolding Phase.} As illustrated in Figure~\ref{fig:teaser}, our key intuition is to replace the frozen off-the-shelf encoder $f$ used by prior latent reasoning methods --- which simply set the latent target to $f(\mathbf{h})$ --- with a \emph{learned} function $g^*\!:\mathbf{h}\mapsto\mathbf{z}^*$ trained directly through the downstream task, so that the resulting $\mathbf{z}^*$ is \emph{optimized for reasoning} rather than tied to a generic vision feature space.

We represent the optimized scaffolding encoder as $g_\psi^*$. It is a trainable copy of the VLM's vision encoder followed by a cross-attention pooling module (Figure~\ref{fig:architecture}, top). It encodes the helper image $\mathbf{h}$ into $K$ latent tokens $\mathbf{z}^* = g_\psi^*(\mathbf{h}) \in \mathbb{R}^{K \times d}$, which are directly inserted at the latent positions of the VLM, after the input $\mathbf{x}$, and the VLM produces the answer $\mathbf{y}$. Given the training input $\mathbf{x}$ (image and text query) and the teacher-forced answer $\mathbf{y}$, we \emph{freeze} the base VLM and update only $\psi$ end-to-end through the standard token-level cross-entropy loss $\mathcal{L}_{\text{CE}}$ on the answer tokens:
\begin{equation}
\mathcal{L}_{\text{scaffolding}}(\psi)
\;=\; \mathcal{L}_{\text{CE}}\!\left(\mathbf{y}\,\big|\,\mathbf{x},\, \mathbf{z}^*\right).
\end{equation}
In this way, the scaffolding encoder learns to produce a $\mathbf{z}^*$ that is trained to capture the reasoning-sensitive features the frozen VLM needs to answer correctly. A trace may contain several helper images, one for each latent block. To encode all of them in one forward pass, we pack them into a single sequence in which each helper image is followed by its $K$ learnable embeddings, $[\mathbf{h}^{(1)},\mathbf{e}^{(1)},\mathbf{h}^{(2)},\mathbf{e}^{(2)},\ldots]$, and apply an attention mask under which each group of embeddings sees only the helper image right before it. The mask keeps the target blocks independent of one another, so they can share a single forward pass (Appendix~\ref{app:attn_mask}).

\paragraph{Latent-Prediction Phase.} Since the helper image $\mathbf{h}$ is unavailable at inference, the VLM is trained to predict its own latent block $\mathbf{z}$ that approaches the optimized target $\mathbf{z}^*$ from the input $\mathbf{x}$ alone (Figure~\ref{fig:architecture}, bottom). Concretely, we freeze the scaffolding encoder $g^*_\psi$ and fine-tune the base VLM with parameters $\theta$. The latent tokens $\mathbf{z}$ are not generated autoregressively; instead, they are generated from $K$ learnable embeddings that pass through the VLM transformer using the input $\mathbf{x}$ as context. This design speeds up inference by producing all $K$ tokens of a block in a single forward pass. During training, we use teacher forcing: each $\langle$BOT$\rangle$ is followed by the $K$ learnable embeddings and then by the target block $\mathbf{z}^*$, giving the sequence $[\mathbf{x},\langle\mathrm{BOT}\rangle,\mathbf{e},\mathbf{z}^*,\langle\mathrm{EOT}\rangle,\mathbf{c},\ldots,\mathbf{y}]$. An attention mask lets the embeddings of a block attend to each other bidirectionally and see all preceding tokens except the embeddings of earlier blocks, and hides them from every other token. Later tokens therefore condition on $\mathbf{z}^*$, and all latent blocks of a trace are predicted in a single forward pass, the \emph{predictor pass} (Appendix~\ref{app:attn_mask}). Since the predictor and the VLM are the same model, a second forward pass then generates the answer text from the latent blocks (Figure~\ref{fig:architecture}, bottom).
We train the base VLM $\theta$ with the following objective:
\begin{equation}
\begin{aligned}
\mathcal{L}_{\text{predict}}(\theta)
  \;=\;& \lambda_{\text{latent}}\, \frac{1}{K}\sum_{k=1}^{K} \left\| \mathbf{z}_{k} - \mathbf{z}^{*}_{k} \right\|_2^2
  \;+\; \lambda_{\text{task}}\, \mathcal{L}_{\text{CE}}\!\left(\mathbf{y}\,\big|\,\mathbf{x},\, \mathbf{z}^{*}\right),
\end{aligned}
\end{equation}
where $\lambda_{\text{latent}}$ and $\lambda_{\text{task}}$ balance the two terms. The first term pulls $\mathbf{z}$ toward the optimized target $\mathbf{z}^*$ from the Scaffolding Phase; the second term trains the VLM to generate the answer from the teacher-forced target $\mathbf{z}^*$, which the predicted $\mathbf{z}$ replaces at inference. With several latent blocks, the first term is averaged over all blocks, and the $\langle$BOT$\rangle$ tokens are included in the cross-entropy loss. At inference, the model decodes text until it emits $\langle$BOT$\rangle$, predicts one latent block in place of $\mathbf{z}^*$, and resumes decoding.

\subsection{Stage 2: Scaffolding RL}
\label{sec:resgrpo}

{Prior latent-reasoning RL methods either optimize only the text trajectory or use a deterministic latent regularizer. Neither choice samples alternative latent actions, so neither directly explores latent reasoning trajectories. Scaffolding RL instead defines an explicit, input-adaptive policy over residual latent actions and samples from it during rollout, as illustrated in Figure~\ref{fig:rl_compare}.
\begin{figure}[!htbp]
  \centering
  \includegraphics[width=0.9\linewidth]{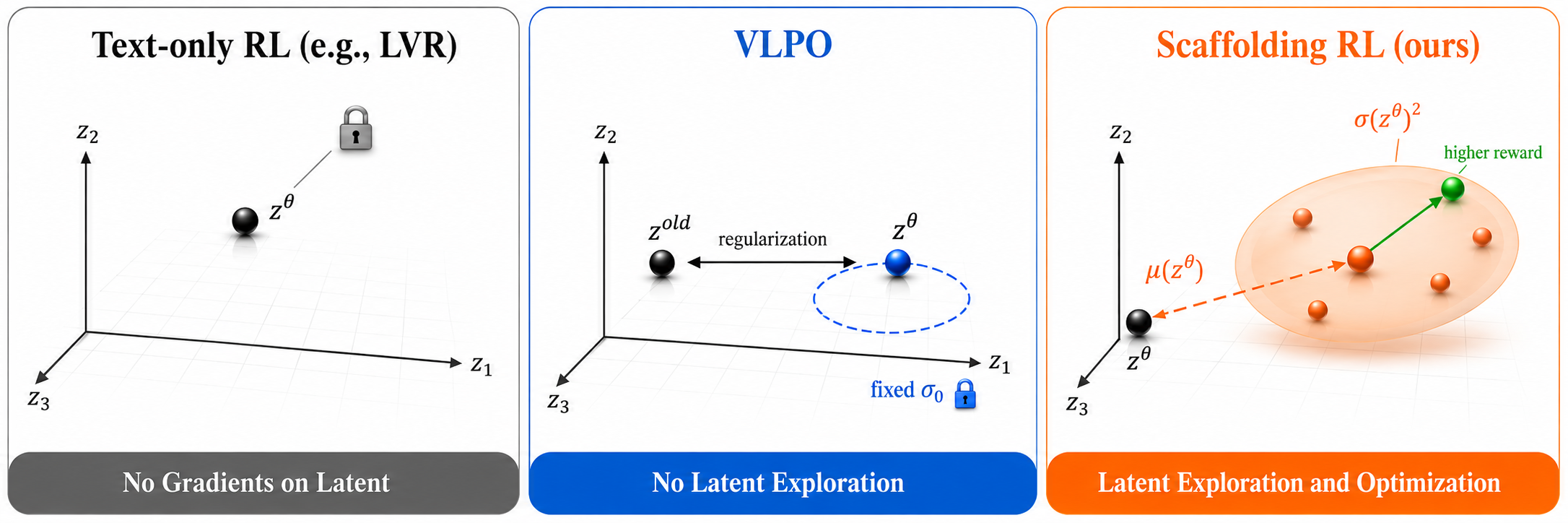}
  \caption{\textbf{Three Stage~2 RL paradigms in latent space.} \textbf{Left:} text-only RL provides no direct gradients on latent actions. \textbf{Middle:} VLPO applies fixed-variance Gaussian regularization to deterministic latent outputs but does not sample latent actions. \textbf{Right:} Scaffolding RL samples residual latent actions from a learned distribution and uses reward to optimize the latent tokens with exploration.}
  \label{fig:rl_compare}
\end{figure}

To solve this, we introduce \textbf{Scaffolding RL}. We treat the generated latent prior $\mathbf{z}^{\theta} = f_{\theta}(\mathbf{x}, \mathbf{c}_{\mathrm{pre}})$ as a \emph{base action} and learn adaptive, reward-guided adjustments $\boldsymbol\Delta_z^*$. Here, $\mathbf{c}_{\mathrm{pre}}$ denotes the optional text reasoning generated before the latent block. Crucially, the mean and variance of the Gaussian sampler are not fixed, but are learned as functions of the base action $\mathbf{z}^{\theta}$. We predict them using two MLP heads on top of the shared VLM hidden states:
\begin{equation}
\boldsymbol{\Delta}_z \sim \mathcal{N}\!\left(\boldsymbol{\mu}(\mathbf{z}^{\theta}),\, \boldsymbol{\sigma}(\mathbf{z}^{\theta})^2\right),
\end{equation}
yielding the perturbed latent block $\mathbf{z}^{\theta} + \boldsymbol{\Delta}_z$ that replaces $\mathbf{z}^{\theta}$ in the forward pass. Because $\boldsymbol{\mu}(\mathbf{z}^{\theta})$ and $\boldsymbol{\sigma}(\mathbf{z}^{\theta})$ are conditioned on the generated latent prior, both the center and scale of exploration can vary across inputs. To ensure stable initial exploration, the mean head is zero-initialized so that the initial distribution is centered on the generated latent prior, while the variance head begins at small positive values. The sampled latent block is treated as one more action of the policy: its probability ratio enters the GRPO objective~\citep{shao2024grpo} alongside those of the text tokens, so the latent block and the text are updated together from the same reward (Appendix~\ref{app:resgrpo}). This dynamic sampling complements the Stage~1 scaffolding encoder and yields the  gains reported in our experiments.


}

\section{Experiments}

\label{sec:experiments}

We evaluate Scaffolding Minds in two complementary settings: \textbf{spatial planning} (Section~\ref{sec:spatial}), requiring multi-step reasoning in structured visual environments, and \textbf{visual-centric reasoning} (Section~\ref{sec:perception}), testing fine-grained evidence extraction and reasoning on real-world benchmarks.


\begin{table}[t]
  \centering
  \caption{\textbf{FrozenLake results (accuracy \%)} on four representative levels; Avg.\ is their mean.  Best in \textbf{bold}, second best \underline{underlined}. Methods marked $^\diamond$ use image-generation architectures. \textcolor{gaingreen}{\textbf{$\Delta$}}: gain of Scaffolding Encoder+RL over SFT. }
  \label{tab:main_results}
  \setlength{\tabcolsep}{5pt}
  \renewcommand{\arraystretch}{1.15}
  \resizebox{\textwidth}{!}{%
  \begin{tabular}{l|cccc|c}
    \toprule
    \rowcolor{headerbg}
    \textbf{Method} & \textbf{Lv.\ 8} {\scriptsize($8{\times}8$)} & \textbf{Lv.\ 16} {\scriptsize($16{\times}16$)} & \textbf{Lv.\ 24} {\scriptsize($24{\times}24$)} & \textbf{Lv.\ 32} {\scriptsize($32{\times}32$)} & \textbf{Avg.} \\
    \midrule
    \rowcolor{basebg}
    \multicolumn{6}{c}{\textit{\small Base model \& supervised fine-tuning}} \\
    SFT & 84.0 & 65.0 & 41.0 & 18.0 & 52.0 \\
    SFT+GRPO & 85.0 & 67.0 & 43.0 & 20.0 & 53.8 \\
    \midrule
    \rowcolor{basebg}
    \multicolumn{6}{c}{\textit{\small Image-generation methods}} \\
    VPRL$^\diamond$~\citep{xu2025} & 88.0 & 62.0 & 48.0 & 35.0 & 58.3 \\
    DiffThinker$^\diamond$~\citep{xu2025diffthinker} & 92.0 & 68.0 & 53.0 & 44.0 & 64.3 \\
    \midrule
    \rowcolor{basebg}
    \multicolumn{6}{c}{\textit{\small Latent reasoning methods}} \\
    LVR~\citep{li2025lvr} & 86.0 & 69.0 & 45.0 & 22.0 & 55.5 \\
    Mirage~\citep{yang2025b} & 87.0 & 71.0 & 47.0 & 24.0 & 57.3 \\
    CoVT~\citep{qin2025covt} & 88.0 & 73.0 & 50.0 & 27.0 & 59.5 \\
    Monet~\citep{wang2025monet} & 89.0 & 75.0 & 53.0 & 31.0 & 62.0 \\
    VaLR~\citep{jeon2026valr} & 91.0 & 78.0 & 57.0 & 36.0 & 65.5 \\
    \midrule
    \rowcolor{oursbg}
    \textbf{Scaffolding Encoder}
    & \underline{94.0}
    & \underline{79.0}
    & \underline{63.0}
    & \underline{52.0}
    & \underline{72.0} \\
    
    \rowcolor{oursbest}
    \textbf{Scaffolding Encoder+RL}
    & \textbf{96.0}
    & \textbf{82.0}
    & \textbf{67.0}
    & \textbf{55.0}
    & \textbf{75.0} \\
    \rowcolor{deltabg}
    \textcolor{gaingreen}{\textbf{$\Delta$ vs.\ SFT}}
        & \textcolor{gaingreen}{\textbf{+12.0}}
        & \textcolor{gaingreen}{\textbf{+17.0}}
        & \textcolor{gaingreen}{\textbf{+26.0}}
        & \textcolor{gaingreen}{\textbf{+37.0}}
        & \textcolor{gaingreen}{\textbf{+23.0}} \\
    \bottomrule
  \end{tabular}}
\end{table}

\paragraph{Implementation.} We use Qwen2.5-VL~\citep{qwen2025vl} as the shared VLM backbone. In Stage~1 (Scaffolding Encoder), we first train the scaffolding encoder---which reuses the VLM vision encoder and compresses the helper image into $K{=}4$ latent tokens---while keeping the remaining VLM components frozen, and then freeze the scaffolding encoder and train the shared VLM to generate the target latent block from the standard input alone. In Stage~2 (Scaffolding RL), we apply reward-guided adjustments on top of the generated latent prior; the scaffolding encoder is no longer used; at inference, the VLM first generates the latent prior, and the two MLP heads then read its last hidden states to produce $\boldsymbol{\mu}$ and $\boldsymbol{\sigma}$, from which the adjustment is sampled.

\paragraph{Baselines.} We compare against three families of methods: (i)~supervised baselines without latent tokens, \emph{SFT} and \emph{SFT+GRPO}; (ii)~latent visual reasoning methods that derive their targets from off-the-shelf vision features --- \emph{LVR}~\citep{li2025lvr}, \emph{Mirage}~\citep{yang2025b}, \emph{CoVT}~\citep{qin2025covt}, \emph{Monet}~\citep{wang2025monet}, and \emph{VaLR}~\citep{jeon2026valr}; and (iii)~image-generation methods that produce explicit intermediate images at inference time, \emph{VPRL}~\citep{xu2025} and \emph{DiffThinker}~\citep{xu2025diffthinker}. On the visual-centric reasoning benchmarks we additionally compare against the thinking-with-images systems \emph{DeepEyes}~\citep{zheng2025deepeyes} and \emph{Thyme}~\citep{zhang2025thyme}, and the latent baseline \emph{SkiLa}~\citep{tong2025skila}.



\subsection{Spatial Planning: FrozenLake}
\label{sec:spatial}
FrozenLake~\citep{wu2024vsp} presents grid-based environments with obstacles (``holes'') requiring multi-step spatial planning: the agent outputs the complete path as directional moves (Up/\allowbreak Down/\allowbreak Left/\allowbreak Right), receiving zero reward for paths through holes. We train a single model on all even levels from 8 to 32, spanning $8{\times}8$ to $32{\times}32$ grids, with 4,550 training examples in total. All grid images are padded to a unified $32{\times}32$ resolution before being encoded. We evaluate on 100 unseen test layouts for each level from 8 to 32, including the odd levels not seen in training, and report four representative levels (8, 16, 24, 32). All baselines are trained and evaluated by us in the same setup; training and evaluation splits are detailed in the appendix.

\begin{wrapfigure}{r}{0.55\textwidth}
  \centering
  \captionsetup{font=small}
  \includegraphics[width=\linewidth]{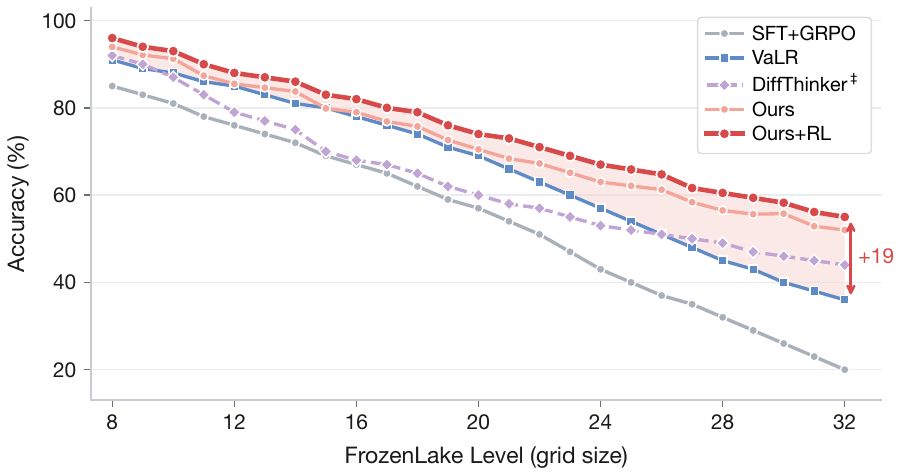}
  \caption{\textbf{Per-level FrozenLake accuracy.} Scaffolding Minds (red) degrades gracefully as grid size grows, while baselines collapse faster. The shaded band highlights the steadily widening advantage of Ours\,+\,RL over the strongest latent baseline (VaLR), reaching \textbf{+19} points at Level~32.}
  \label{fig:main_results}
\end{wrapfigure}

\paragraph{Main results.} Table~\ref{tab:main_results} and Figure~\ref{fig:main_results} show a consistent pattern across maze complexity. First, compared with base methods, our full model improves average accuracy by 23.0 points over the SFT baseline, indicating that latent visual reasoning remains valuable even on top of a strong base. Second, compared with latent reasoning baselines, our method gains 9.5 points over the strongest prior approach. This gap is important because the latent baselines are trained with the same intermediate helper images; the advantage therefore comes from learning a better latent target space rather than from access to extra information. Our latent representation is optimized directly through downstream task loss, which makes it more suitable for decision-relevant reasoning than targets inherited from general-purpose vision features. Third, compared with image-generation methods, our method improves by 10.7 points. This suggests that explicitly generating intermediate images is not necessary to obtain strong visual reasoning, as long as the latent space itself captures the right reasoning structure. Figure~\ref{fig:main_results} further shows that this advantage grows with difficulty: our method degrades more gracefully as the maze becomes larger, reaching a +19-point gap over the strongest latent baseline at Level~32.

\begin{table}[t]
  \centering
  \caption{\textbf{Visual-centric reasoning benchmark results (accuracy \%).} \textbf{Bold}: best; \underline{underlined}: second best. $^\dag$: reproduced via original checkpoint/code; $^\ddagger$: our re-implementation. $^\S$: retrained by us under our setup. Unmarked baseline cells are taken from the original papers.  CVB.=CVBench, HRB=HRBench, MME=MME-RealWorld-Lite, Jig.=Jigsaw.}
  \label{tab:perception}
  \setlength{\tabcolsep}{4.2pt}
  \renewcommand{\arraystretch}{1.15}
  \resizebox{\textwidth}{!}{%
  \begin{tabular}{l|ccccccccc|c}
    \toprule
    \rowcolor{headerbg}
    \textbf{Method} & \textbf{V$^\star$} & \textbf{BLINK} & \textbf{MMVP} & \textbf{MMStar} & \textbf{CVB.} & \textbf{HRB-4K} & \textbf{HRB-8K} & \textbf{MME} & \textbf{Jig.} & \textbf{Avg.} \\
    \midrule
    \rowcolor{basebg}
    \multicolumn{11}{c}{\textit{\small Closed-source models}} \\
    GPT-4o & 66.0 & 60.0 & 70.7 & 61.6 & 80.1 & 59.0 & 55.5 & 52.0 & 53.0 & 62.0 \\
    GPT-4V & 58.0 & 58.3 & 51.0 & 56.0 & 69.5 & 56.5 & 52.0 & 45.0 & 47.0 & 54.8 \\
    GPT-4o-mini & 57.0 & 53.6 & 56.0 & 54.8 & 68.0 & 56.0 & 51.0 & 37.4 & 46.0 & 53.3 \\
    Claude\,3.7-Sonnet & 72.0 & 56.6 & 64.0 & 65.1 & 80.0 & 68.0 & 64.0 & 49.0 & 52.0 & 63.4 \\
    \midrule
    \rowcolor{basebg}
    \multicolumn{11}{c}{\textit{\small Base model \& supervised fine-tuning}} \\
    Qwen2.5-VL-7B & 76.2 & 55.7 & 56.0 & 63.9 & 74.5 & 68.6 & 64.9 & 39.7 & 50.4 & 61.1 \\
    Vanilla SFT & 81.3 & 57.2 & 58.7 & 66.0 & 77.0 & 70.5 & 66.8 & 42.0 & 52.5 & 63.6 \\
    \midrule
    \rowcolor{basebg}
    \multicolumn{11}{c}{\textit{\small Tool-calling \& think-with-image methods}} \\
    DeepEyes$^\S$~\citep{zheng2025deepeyes} & 83.5 & 59.5 & 61.7 & 67.8 & 80.6 & 75.3 & \underline{72.8} & 53.4 & 54.6 & 67.7 \\
    Thyme~\citep{zhang2025thyme} & 81.9 & 56.1 & 62.4$^\dag$ & 65.9 & 79.8$^\dag$ & \underline{77.0} & 72.0 & 55.2 & 54.8$^\dag$ & 67.2 \\
    \midrule
    \rowcolor{basebg}
    \multicolumn{11}{c}{\textit{\small Latent reasoning methods}} \\
    Mirage~\citep{yang2025b} & 80.7$^\ddagger$ & 57.4$^\ddagger$ & 57.6$^\ddagger$ & 65.7$^\ddagger$ & 76.3$^\ddagger$ & 69.8$^\ddagger$ & 66.4$^\ddagger$ & 42.3$^\ddagger$ & 53.6$^\ddagger$ & 63.3 \\
    CoVT~\citep{qin2025covt} & 78.0 & 56.0 & 58.7 & 68.9$^\dag$ & 80.0 & 72.9 & 69.4 & 51.4$^\dag$ & 52.7$^\dag$ & 65.3 \\
    Monet~\citep{wang2025monet} & 83.2 & 57.7$^\dag$ & 60.3$^\dag$ & 68.3$^\dag$ & 79.7$^\dag$ & 71.0 & 68.0 & \underline{55.5} & 54.6$^\dag$ & 66.5 \\
    LVR~\citep{li2025lvr} & 81.7 & 56.7$^\dag$ & 71.7 & 67.2$^\dag$ & 78.3$^\dag$ & 71.4$^\dag$ & 67.6$^\dag$ & 44.3$^\dag$ & 52.4$^\dag$ & 65.7 \\
    VaLR$^\S$~\citep{jeon2026valr}
    & 84.6
    & 62.3
    & 70.2
    & 69.7
    & 83.2
    & 70.6
    & 68.7
    & 50.1
    & 53.8
    & 68.1 \\
    SkiLa$^\S$~\citep{tong2025skila} & 84.8 & 57.5 & \underline{75.7} & 65.4 & 82.1 & 72.8 & 67.3 & 54.6 & 54.3 & 68.3 \\
    \midrule
    \rowcolor{oursbg}
    \textbf{Scaffolding Encoder} & \underline{87.4} & \underline{64.8} & 73.3 & \underline{72.1} & \underline{86.4} & 73.6 & 71.1 & 53.4 & \underline{56.0} & \underline{70.9} \\
    \rowcolor{oursbest}
    \textbf{Scaffolding Encoder+RL} & \textbf{90.6} & \textbf{67.2} & \textbf{76.7} & \textbf{73.5} & \textbf{88.4} & \textbf{77.5} & \textbf{74.1} & \textbf{57.3} & \textbf{59.7} & \textbf{73.9} \\
    \midrule
    \rowcolor{deltabg}
    \textcolor{gaingreen}{\textbf{$\Delta$ vs.\ Qwen2.5-VL-7B}}
        & \textcolor{gaingreen}{\textbf{+14.4}}
        & \textcolor{gaingreen}{\textbf{+11.5}}
        & \textcolor{gaingreen}{\textbf{+20.7}}
        & \textcolor{gaingreen}{\textbf{+9.6}}
        & \textcolor{gaingreen}{\textbf{+13.9}}
        & \textcolor{gaingreen}{\textbf{+8.9}}
        & \textcolor{gaingreen}{\textbf{+9.2}}
        & \textcolor{gaingreen}{\textbf{+17.6}}
        & \textcolor{gaingreen}{\textbf{+9.3}}
        & \textcolor{gaingreen}{\textbf{+12.8}} \\
    \bottomrule
  \end{tabular}}
\end{table}

\subsection{Visual-Centric Reasoning Benchmarks}
\label{sec:perception}
To show that Scaffolding Minds generalizes beyond spatial planning and is not tied to a task-specific helper image such as the FrozenLake heatmap, we evaluate on nine established visual-centric reasoning benchmarks: V$^\star$~\citep{wu2024vstar} (fine-grained attribute/spatial reasoning), BLINK~\citep{fu2024blink} (core perception), MMVP~\citep{tong2024mmvp} (visual shortcomings), MMStar~\citep{chen2024mmstar} (vision-indispensable), CVBench~\citep{tong2024cvbench} (2D/3D understanding), HRBench-4K and HRBench-8K~\citep{wang2024hrbench} (high-resolution), MME-RealWorld-Lite~\citep{zhang2025mmerealworld} (real-world scenarios), and Jigsaw~\citep{lyu2025jigsaw} (spatial reasoning). We train on Zebra-CoT~\citep{zebracot}, using 162K interleaved traces for Stage~1 and a disjoint 20K split for Stage~2. Each trace alternates text with a variable number of dataset-provided helper images, such as crops, bounding boxes, diagrams, and candidate placements. These helpers provide intermediate visual evidence rather than the solution, and each supervises one latent block, so the model learns to produce multiple interleaved blocks. Almost all key settings follow FrozenLake (Appendices~\ref{app:visual_setup} and~\ref{app:vision_helpers}).
\paragraph{Results.} Table~\ref{tab:perception} and Figure~\ref{fig:teaser} (right) show that the same latent-learning strategy transfers beyond FrozenLake. First, compared with the supervised fine-tuning baseline, our method improves the average score by 10.3 points, showing that the learned latent space remains useful even on visual-centric reasoning benchmarks without explicit planning outputs. Second, compared with prior latent reasoning methods, our method improves the strongest prior average by 5.6 points. The improvement is consistent across benchmarks that require spatial discrimination and fine-grained visual comparison, where our method also gains the most over the base model, e.g., $+20.7$ points on MMVP, $+17.6$ points on MME-RealWorld-Lite, and $+14.4$ points on V$^\star$.  Third, compared with prior thinking-with-images style methods, our method improves the strongest prior average by 6.2 points, indicating that strong visual-reasoning gains do not require manipulating or generating additional images at inference time when the latent representation is already optimized for the downstream task.  


\Needspace{0.32\textheight}
\subsection{Ablation on Key Contributions}
\label{sec:ablation}

\begin{wrapfigure}{r}{0.50\textwidth}
  \centering
  \captionsetup{font=small}
  \includegraphics[width=\linewidth]{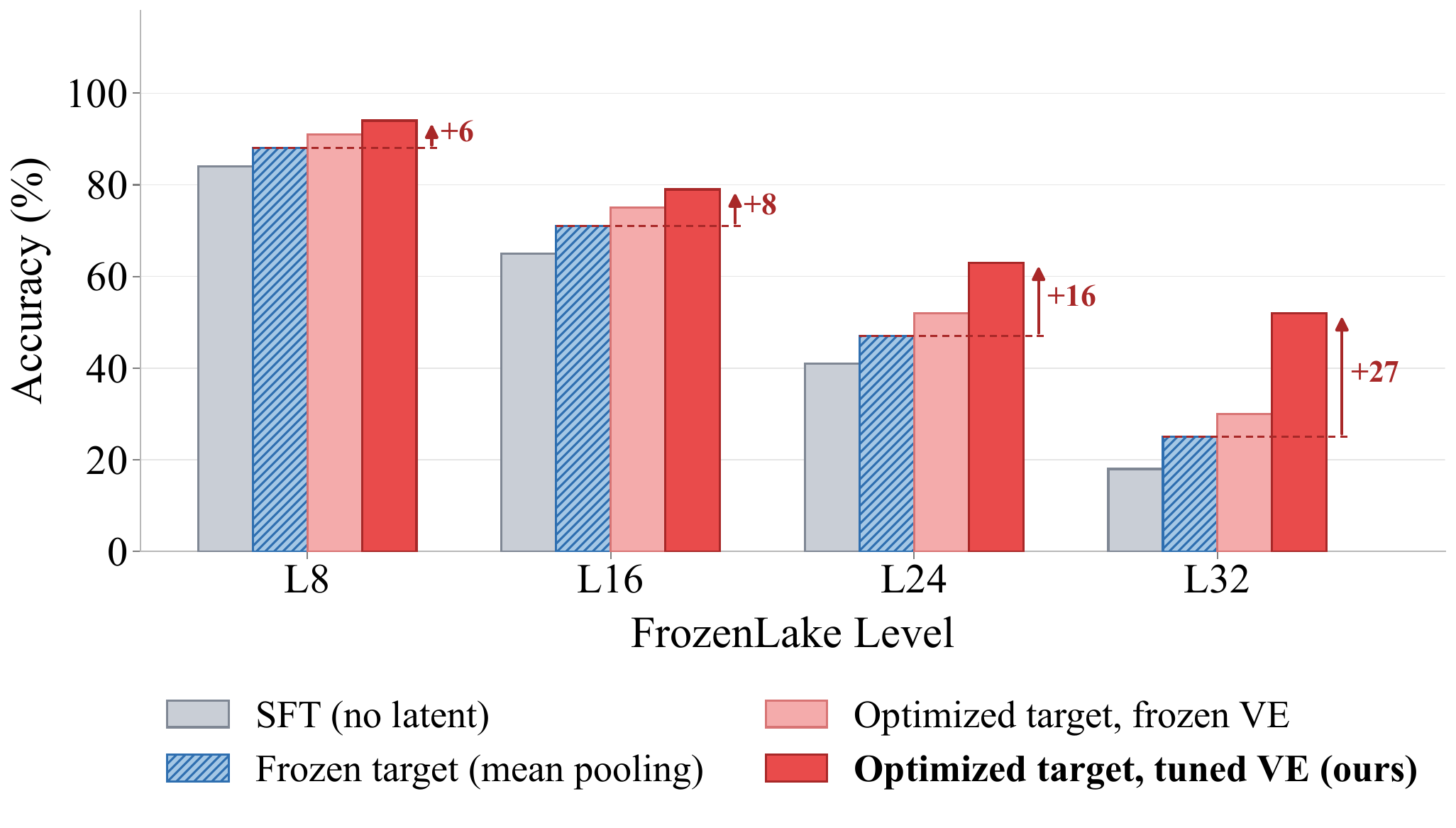}
  \caption{Stage 1 visual target ablation on FrozenLake. } 
  \label{fig:ablation}
\end{wrapfigure}

\paragraph{SFT visual target ablation.} To identify which Stage~1 design choice drives the gain over prior latent reasoning methods, we compare four Stage~1 configurations under the same Qwen2.5-VL-7B backbone with no Stage~2 RL applied: \emph{SFT} without latent tokens as a control; \emph{Frozen target}, the prior-work objective that L2-regresses latent tokens onto frozen off-the-shelf vision-encoder features~\citep{li2025lvr}, mean-pooled to $K$ tokens so that the target has no trainable parameters; and two \emph{optimized targets}, trained through the downstream task loss, that differ only in how much of the target encoder is trained: the attention-pooling module alone (${\sim}80$M parameters; \emph{Optimized target, frozen VE}) or also the vision encoder (${\sim}0.76$B in total; \emph{Optimized target, tuned VE}, ours). Attention pooling is simply the smallest trainable module through which a target can be optimized. As shown in Figure~\ref{fig:ablation}, optimizing the target lifts accuracy by $+4.3$ points even when only the pooling module is trained, and letting the task loss also shape the vision encoder adds a further $+10.0$ points; together they account for the full $+14.3$-point gap over the frozen target used by prior work. Added capacity alone does not explain this: tuning the same vision encoder in plain SFT gains only $+4.5$ points (Appendix~\ref{app:more_ablations}). \emph{These results suggest that optimizing the latent target, rather than regressing onto frozen features, drives the Stage~1 gain.}

\begin{table}[t]
\centering
\captionsetup{font=normalsize}
\caption{\textbf{Stage~2 RL ablation on FrozenLake.} We report the mean and standard deviation of accuracy over 3 seeds, averaged over the four representative Levels~8, 16, 24, and 32.}
\label{tab:rl_ablation}

\normalsize
\setlength{\tabcolsep}{5pt}
\renewcommand{\arraystretch}{1.08}

\begin{tabular}{l|cc|cc}
\toprule
\rowcolor{headerbg}
& \multicolumn{2}{c|}{\textbf{Latent sampling}}
& \multicolumn{2}{c}{\textbf{Stage~1 target}} \\

\rowcolor{headerbg}
\textbf{Stage~2 recipe}
& \textbf{Mean}
& \textbf{Variance}
& \textbf{Scaffolding enc.}
& \textbf{Frozen target} \\

\midrule

\rowcolor{gray!15}
\textit{Stage~1 checkpoint (no RL)}
& \multicolumn{2}{c|}{\textit{--}}
& \textit{72.0}{\scriptsize$\pm0.6$}
& \textit{57.8}{\scriptsize$\pm0.6$} \\

\midrule

Text-only GRPO
& \multicolumn{2}{c|}{no latent sampling}
& 72.6{\scriptsize$\pm0.5$} ($+0.6$)
& 58.4{\scriptsize$\pm0.5$} ($+0.6$) \\

VLPO~\citep{wang2025monet}
& \multicolumn{2}{c|}{latent regularization}
& 72.9{\scriptsize$\pm0.6$} ($+0.9$)
& 59.0{\scriptsize$\pm0.6$} ($+1.2$) \\

\midrule

Learned mean only
& learned
& fixed
& 73.6{\scriptsize$\pm0.5$} ($+1.6$)
& 60.4{\scriptsize$\pm0.7$} ($+2.6$) \\

Learned variance only
& zero
& learned
& 73.2{\scriptsize$\pm0.6$} ($+1.2$)
& 59.8{\scriptsize$\pm0.5$} ($+2.0$) \\

\rowcolor{oursbg}
\textbf{Scaffolding RL (ours)}
& \textbf{learned}
& \textbf{learned}
& \textbf{75.0}{\scriptsize$\pm0.7$} ($\mathbf{+3.0}$)
& \textbf{63.0}{\scriptsize$\pm0.6$} ($\mathbf{+5.2}$) \\

\bottomrule
\end{tabular}

\end{table}

\paragraph{RL method comparison.} Table~\ref{tab:rl_ablation} compares Stage~2 recipes from the same Stage~1 checkpoint. Text-only GRPO leaves the latent block untouched and adds only $+0.6$ points. VLPO adds $+0.9$ points, but its training fluctuates (Figure~\ref{fig:rl_full}) because its fixed-variance Gaussian regularizes the latent block without sampling alternative latent rollouts. Scaffolding RL gives the largest gain ($+3.0$ points), and its advantage widens on the frozen-target checkpoint ($+5.2$ vs.\ $+1.2$ points for VLPO), where the weaker latent leaves more room for latent exploration. Neither head suffices alone: a learned mean or a learned variance gains only $+1.6$ or $+1.2$ points, since the mean sets the direction of exploration and the variance sets its width. \emph{Sampling learned adjustments on top of the Stage~1 prior is essential for converting reward into reliable latent-space exploration.}

\long\gdef\postponedablations{%
\subsection{Ablation on Hyperparameters}
\label{sec:analysis}

\begin{table}[H]
\centering
\begin{minipage}[t]{0.40\linewidth}
\centering
\caption{\textbf{Number of latent tokens $K$} on FrozenLake (Stage~1 only, accuracy \%).}
\label{tab:hyper_K}
\setlength{\tabcolsep}{4.8pt}
\renewcommand{\arraystretch}{1.1}
\resizebox{\linewidth}{!}{%
\begin{tabular}{c|cccc|c}
  \toprule
  \rowcolor{headerbg}
  $K$ & \textbf{L8} & \textbf{L16} & \textbf{L24} & \textbf{L32} & \textbf{Avg.} \\
  \midrule
  1 & 85.0 & 70.0 & 54.0 & 45.0 & 63.5 \\
  2 & 88.0 & 74.0 & 57.0 & 48.0 & 66.8 \\
  \textbf{4} & \textbf{94.0} & \textbf{79.0} & \textbf{63.0} & \textbf{52.0} & \textbf{72.0} \\
  8 & 93.0 & 78.0 & 61.0 & 51.0 & 70.8 \\
  16 & 91.0 & 76.0 & 59.0 & 50.0 & 69.0 \\
  \bottomrule
  \end{tabular}}
\end{minipage}\hfill
\begin{minipage}[t]{0.56\linewidth}
\centering
\caption{\textbf{Helper image choice on FrozenLake} (Stage~1 only, accuracy \%). \emph{Scaffolding encoder}~$\checkmark$: our scaffolding encoder is used; $\times$: a frozen off-the-shelf vision encoder is used as in prior work.}
\label{tab:helper_choice}
\setlength{\tabcolsep}{4.5pt}
\renewcommand{\arraystretch}{1.1}
\resizebox{\linewidth}{!}{%
\begin{tabular}{l|c|cccc|c}
\toprule
\rowcolor{headerbg}
\textbf{Helper image} & \textbf{Scaffold. enc.} & \textbf{L8} & \textbf{L16} & \textbf{L24} & \textbf{L32} & \textbf{Avg.} \\
\midrule
\multirow{2}{*}{Red-arrow overlay} & $\times$ & 85.0 & 67.0 & 52.0 & 40.0 & 61.0 \\
                                        & $\checkmark$ & 92.0 & 76.0 & 58.0 & 46.0 & 68.0 \\
\midrule
\multirow{2}{*}{Value-function heatmap} & $\times$ & 88.0 & 71.0 & 47.0 & 25.0 & 57.8 \\
                                                   & $\checkmark$ & \textbf{94.0} & \textbf{79.0} & \textbf{63.0} & \textbf{52.0} & \textbf{72.0} \\
\bottomrule
\end{tabular}}
\end{minipage}
\end{table}

\paragraph{Number of latent tokens $K$.} We sweep $K \in \{1, 2, 4, 8, 16\}$ on FrozenLake (Levels~8, 16, 24, and 32) while keeping all other Stage~1 settings fixed (Table~\ref{tab:hyper_K}). Accuracy rises sharply from $K{=}1$ ($63.5\%$) to $K{=}4$ ($72.0\%$), then falls back at $K{=}8$ ($70.8\%$) and $K{=}16$ ($69.0\%$); $K{=}4$ is the best setting at every level. This non-monotonic pattern reflects two opposing pressures: too few tokens cannot encode a multi-step spatial plan, while too many tokens add redundancy that latent generation must learn to match. $K{=}4$ strikes the balance between expressiveness and learnability.

\paragraph{Helper image choice.} We cross two helper images on FrozenLake --- the dense \emph{value-function heatmap} (our default) and a sparse \emph{red-arrow overlay} that draws the optimal path on the raw grid --- with the choice between our scaffolding encoder and a frozen off-the-shelf encoder (Table~\ref{tab:helper_choice}, $2{\times}2$). \textbf{First}, a frozen off-the-shelf encoder cannot exploit a richer helper image: with this encoder, the dense heatmap is actually worse than the sparse red-arrow overlay ($57.8$ vs.\ $61.0$), because an encoder trained on natural images cannot readily use the non-natural heatmap. \textbf{Second}, the scaffolding encoder fixes this. It beats the off-the-shelf encoder for both helpers, with the largest gain on the non-natural heatmap ($+14.3$ points, $57.8 \to 72.0$) versus only $+7.0$ points on the red-arrow overlay ($61.0 \to 68.0$); the ordering between helpers also flips, with the heatmap now beating the red-arrow overlay ($72.0$ vs.\ $68.0$).
}

\section{Related Work}
\label{sec:related}

\paragraph{Think about Image.}
This line of work treats the image as a static precondition and reasons \emph{about} it purely in language space, extending textual chain-of-thought~\citep{wei2022, kojima2022zeroshot} to vision-language models such as Qwen2.5-VL~\citep{qwen2025vl}, LLaVA~\citep{liu2024llava}, and Gemini~\citep{team2023gemini} by verbalizing the visual evidence into a natural-language reasoning chain~\citep{fuyu2023, xu2024, zhao2024, dong2025insight, mitra2024, mondal2024, shao2024a, cheng2025, wei2025, du2025gcot}. \emph{Unlike these text-only CoTs, which tend to lose fine-grained spatial and perceptual evidence, our method carries such evidence forward as continuous latent visual tokens in the reasoning chain.} A second line of work, \emph{thinking with images}, instead manipulates or generates intermediate images during reasoning; we discuss it in Appendix~\ref{app:related_twi}.
\paragraph{Latent Reasoning.}
A more recent line replaces explicit intermediate images with a compact block of continuous latent tokens inserted into the reasoning chain, building on observations that explicit chain-of-thought has its own limitations~\citep{pinker1994, silver2016, chen2024melodi}. On the text side, related work has explored pause and planning tokens~\citep{goyal2024, wang2024}, compressed CoT~\citep{cheng2024durme, shen2025}, continuous reasoning spaces~\citep{hao2024, liu2024deliberation}, latent internalization of CoT~\citep{deng2023, deng2024, zelikman2022}, latent diffusion~\citep{kang2025ladir, kang2026ladirl}, and looped architectures~\citep{saunshi2025, teoh2025nextlat}. In the multimodal setting, prior methods derive the latent target from a frozen off-the-shelf vision encoder (e.g., SigLIP~\citep{tschannen2025siglip}) applied to a helper image, through variants including L2 feature regression, vision-aligned distillation, policy-optimized embeddings, and sketchpad-style decoders~\citep{yang2025b, wang2025monet, qin2025covt, li2025lvr, tong2025skila, zhang2025sketchpad, chen2025, jeon2026valr, dong2026interleaved}. Methodologically, this family is closely related to joint-embedding predictive architectures~\citep{lecun2022jepa} and learning-using-privileged-information~\citep{vapnik2009lupi}, and conceptually echoes the idea of fading instructional scaffolds in cognitive science~\citep{vygotsky1978, wood1976scaffolding}. \emph{In contrast to prior multimodal latent methods that inherit a fixed general-purpose target from a frozen vision encoder, our method trains a scaffolding encoder end-to-end through the downstream task loss --- turning the latent target itself into an optimized component of the system --- and pairs it with an input-adaptive RL sampler whose mean and variance are learned functions of the latent prior.}

\section{Conclusion}

\label{sec:conclusion}

We present Scaffolding Minds, a two-stage framework that improves both how latent visual reasoning is supervised and how it is refined through RL. Stage~1 learns a scaffolding encoder whose latent target is optimized end-to-end for the reasoning task, in place of the fixed vision-encoder target used by prior work. Stage~2 learns a Gaussian latent policy that samples residual actions around the predicted latent, enabling reward-driven exploration directly in latent space. Together, the two stages improve the strongest prior latent baseline by $+9.5$ points on average on FrozenLake ($+19$ points at $32{\times}32$) and by $+5.6$ points on average across nine visual-centric reasoning benchmarks, showing that both the quality of the latent target and direct latent-space exploration matter.

\bibliographystyle{abbrvnat}
\bibliography{references}

\appendix
\newpage

\section{Limitation and Future Work}
\label{app:limitations}

In line with previous latent reasoning methods, our approach relies on the availability of an intermediate helper image during training; how to extend the framework to settings where such helper images are unavailable or expensive to obtain remains an open question. In addition, while we use value-function heatmaps as the natural helper-image choice for FrozenLake, we do not exhaustively search alternative helper-image designs, and automatically proposing helper images for new domains is left to future work. Finally, each latent block uses a fixed number of $K$ tokens, and adapting the block length to the difficulty of each reasoning step is a natural extension.

\section{Additional Related Work}
\label{app:related_twi}

\paragraph{Think with Image.}
A second line of work lets the model actively manipulate or generate intermediate images \emph{during} reasoning. Tool-augmented systems call external tools or code to crop, zoom, or annotate the input~\citep{yao2023, gao2023, hu2024b, yang2023}; image-generation systems autoregressively visualize new images~\citep{li2025mvot, gu2025thinkmorph, xu2025, yang2025, zhang2025thyme} or use diffusion-based generation~\citep{xu2025diffthinker}; and RL-driven variants train the VLM to decide when to invoke a tool or emit a new image~\citep{su2025, zheng2025deepeyes, meng2025mmeureka}. These methods preserve visual evidence across reasoning steps but pay a substantial inference-time cost from explicit image manipulation or generation. \emph{In contrast, our method retains a similar multi-step visual reasoning capability while operating entirely in latent space, at near-base-VLM inference cost.}


\section{Scaffolding RL: Detailed Derivation}
\label{app:resgrpo}

This appendix provides the full mathematical derivation of the Scaffolding RL objective summarized in Section~\ref{sec:resgrpo}.

\paragraph{Setup.} Let $\mathbf{x}$ denote the model input (image and query), and let $\mathbf{z}^{\mathrm{old}}=f_{\theta_{\mathrm{old}}}(\mathbf{x})$ and $\mathbf{z}^{\theta}=f_{\theta}(\mathbf{x})$ denote the Stage~1 latent priors produced by the rollout and current policies. Their Gaussian heads produce $\boldsymbol{\mu}^{\mathrm{old}}_k,\boldsymbol{\sigma}^{\mathrm{old}}_k$ and $\boldsymbol{\mu}^{\theta}_k,\boldsymbol{\sigma}^{\theta}_k$, respectively. For rollout $i$, the old latent policy samples a residual adjustment at each latent position $k$:
\begin{equation}
  \boldsymbol{\delta}^{\mathrm{old}}_{i,k} \sim \mathcal{N}\!\left(\boldsymbol{\mu}^{\mathrm{old}}_{k},\, (\boldsymbol{\sigma}^{\mathrm{old}}_{k})^2\right),
  \qquad
  \mathbf{a}^{\mathrm{old}}_{i,k} = \mathbf{z}^{\mathrm{old}}_k + \boldsymbol{\delta}^{\mathrm{old}}_{i,k}.
\end{equation}
Stacking all positions gives the rollout latent action $\mathbf{a}^{\mathrm{old}}_i=[\mathbf{a}^{\mathrm{old}}_{i,1},\ldots,\mathbf{a}^{\mathrm{old}}_{i,K}]$. Conditioned on $\mathbf{x}$ and $\mathbf{a}^{\mathrm{old}}_i$, the model samples a text answer $y_i\sim\pi_{\theta_{\mathrm{old}}}(\cdot\mid\mathbf{x},\mathbf{a}^{\mathrm{old}}_i)$ of length $T_i$.

\paragraph{Block-level latent importance ratio.} We evaluate the same sampled latent action under the rollout and current factorized diagonal Gaussians:
\begin{equation}
  p_{\vartheta}(\mathbf{a}^{\mathrm{old}}_i\mid\mathbf{x})=
  \prod_{k=1}^{K}\prod_{j=1}^{d}
  \frac{1}{\sqrt{2\pi}\,\sigma^{\vartheta}_{k,j}}
  \exp\!\left[-\frac{1}{2}
  \left(\frac{a^{\mathrm{old}}_{i,k,j}-z^{\vartheta}_{k,j}-\mu^{\vartheta}_{k,j}}
  {\sigma^{\vartheta}_{k,j}}\right)^2\right],
  \qquad \vartheta\in\{\theta,\theta_{\mathrm{old}}\},
\end{equation}
where $j$ indexes the $d$ latent dimensions. The block-level latent importance ratio is
$\rho^{\mathrm{lat}}_{i}(\theta)=p_{\theta}(\mathbf{a}^{\mathrm{old}}_i\mid\mathbf{x})\,/\,p_{\theta_{\mathrm{old}}}(\mathbf{a}^{\mathrm{old}}_i\mid\mathbf{x})$. If a rollout contains several latent blocks, the ratio is computed for each block, and the latent term below is averaged over the blocks.

\paragraph{Token-level text importance ratio.} For the discrete answer trajectory we use the standard per-token ratio:
$\rho^{\mathrm{text}}_{i,t}(\theta)=\pi_{\theta}(y_{i,t}\mid y_{i,<t},\mathbf{x},\mathbf{a}^{\mathrm{old}}_i)\,/\,\pi_{\theta_{\mathrm{old}}}(y_{i,t}\mid y_{i,<t},\mathbf{x},\mathbf{a}^{\mathrm{old}}_i)$.

\paragraph{Combined clipped objective.} Scaffolding RL optimizes a GRPO-style clipped objective that sums the latent term and the text term:
\begin{align}
  \mathcal{L}_{\mathrm{RL}}
  = -\frac{1}{G}\sum_{i=1}^{G} \Bigg[
  &\min\!\Big(
    \rho^{\mathrm{lat}}_{i}(\theta) A_i,\,
    \operatorname{clip}\!\big(\rho^{\mathrm{lat}}_{i}(\theta), 1{-}\epsilon, 1{+}\epsilon\big) A_i
  \Big)
  \nonumber\\
  &+
  \frac{1}{T_i}\sum_{t=1}^{T_i}
  \min\!\Big(
    \rho^{\mathrm{text}}_{i,t}(\theta) A_i,\,
    \operatorname{clip}\!\big(\rho^{\mathrm{text}}_{i,t}(\theta), 1{-}\epsilon, 1{+}\epsilon\big) A_i
  \Big)
  \Bigg]
  \nonumber\\
  &\quad+\frac{\beta_{\mathrm{KL}}}{G}\sum_{i=1}^{G}\frac{1}{T_i}\sum_{t=1}^{T_i}
  D_{\mathrm{KL}}\!\left(
  \pi_{\theta}(\cdot\mid s_{i,t})\,\|\,
  \pi_{\mathrm{ref}}(\cdot\mid s_{i,t})
  \right),
\end{align}
where $\epsilon$ is the GRPO clipping threshold, $G$ is the rollout group size, $A_i$ is the group-relative advantage, $s_{i,t}=(y_{i,<t},\mathbf{x},\mathbf{a}^{\mathrm{old}}_i)$ is the token-generation state, $\pi_{\mathrm{ref}}$ is the fixed Stage~1 reference policy, and $\beta_{\mathrm{KL}}$ is the KL coefficient. Since the $K$ latent tokens are predicted jointly, the latent term acts on the sampled latent block as a whole rather than on individual autoregressive steps. During Stage~2, $f_{\theta}$ and both Gaussian heads remain trainable; no gradient is stopped or detached through $\mathbf{z}^{\theta}$.

\section{Experimental Setup Details}
\label{app:additional_experimental_details}

\subsection{FrozenLake}

\paragraph{Data.} We use the FrozenLake suite from VSP~\citep{wu2024vsp}. Training data covers even levels from 8 to 32 ($8{\times}8$ to $32{\times}32$ grids), with the per-level training set growing from 50 examples at Level~8 to 650 at Level~32 in steps of 50, for a total of 4{,}550 training examples. Each level has a held-out test set of 100 examples that we use for evaluation. Each of the four representative levels (8, 16, 24, 32) also has a held-out validation set of 50 examples (half the size of the test set), disjoint from both the training and test sets; we select $K$, the loss weights $(\lambda_{\mathrm{latent}},\lambda_{\mathrm{task}})$, and the helper image on this validation set, and all reported numbers are test accuracy. All grid images are padded to a unified $32{\times}32$ resolution before being passed to the vision encoder, so that token counts and positional encodings are constant across difficulty levels. The four representative levels reported in Table~\ref{tab:main_results} (8, 16, 24, 32) are chosen to span the full difficulty range.

\paragraph{Backbone and architecture.} Unless noted otherwise, all results use Qwen2.5-VL-7B~\citep{qwen2025vl} as the shared VLM backbone. The scaffolding encoder is implemented as a trainable copy of the VLM's vision encoder followed by a cross-attention pooling module that compresses the resulting visual tokens into $K{=}4$ latent tokens. Latent generation reuses the full VLM weights with $K{=}4$ learnable embeddings that attend to the shared representation to produce the latent block in a single forward pass. In Stage~2, the mean and variance heads are two-layer MLPs operating on the shared VLM hidden states at the latent positions.

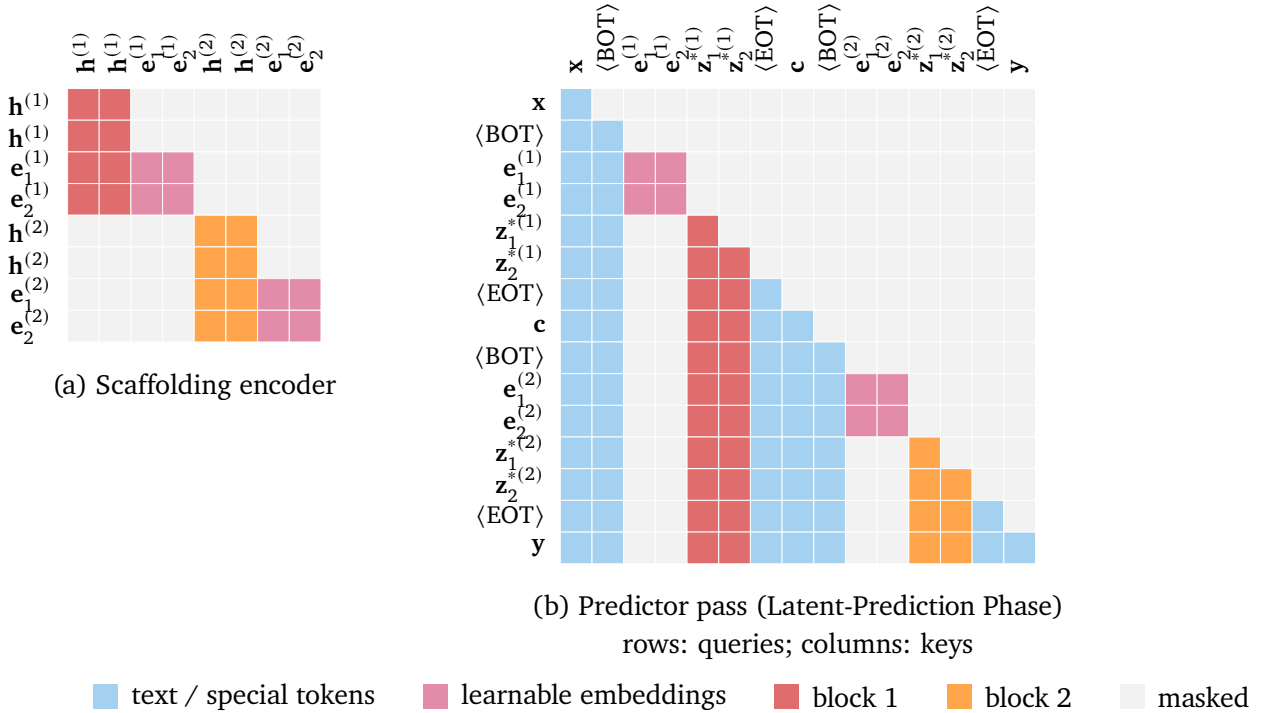
\begin{figure}[!htbp]
  \centering
  \resizebox{\linewidth}{!}{%
  \begin{tikzpicture}[x=0.36cm,y=0.36cm,font=\scriptsize]
    \def\atype{{1,1,2,2,3,3,4,4}}
    \foreach \lab [count=\n from 0] in {$\mathbf{h}^{(1)}$,$\mathbf{h}^{(1)}$,$\mathbf{e}^{(1)}_{1}$,$\mathbf{e}^{(1)}_{2}$,$\mathbf{h}^{(2)}$,$\mathbf{h}^{(2)}$,$\mathbf{e}^{(2)}_{1}$,$\mathbf{e}^{(2)}_{2}$}{
      \node[anchor=east] at (-0.1,-\n-0.5) {\lab};
      \node[anchor=west,rotate=90] at (\n+0.5,0.1) {\lab};
    }
    \foreach \i in {0,...,7}{
      \foreach \j in {0,...,7}{
        \pgfmathtruncatemacro{\ti}{\atype[\i]}
        \pgfmathtruncatemacro{\tj}{\atype[\j]}
        \pgfmathtruncatemacro{\ok}{(\ti==1 && \tj==1) || (\ti==2 && (\tj==1 || \tj==2)) || (\ti==3 && \tj==3) || (\ti==4 && (\tj==3 || \tj==4)) ? 1 : 0}
        \ifnum\ok=1
          \ifnum\tj=1 \fill[oursred!70] (\j,-\i) rectangle ++(1,-1);
          \else\ifnum\tj=3 \fill[orange!70] (\j,-\i) rectangle ++(1,-1);
          \else \fill[purple!45] (\j,-\i) rectangle ++(1,-1); \fi\fi
        \else
          \fill[gray!10] (\j,-\i) rectangle ++(1,-1);
        \fi
        \draw[white,line width=0.3pt] (\j,-\i) rectangle ++(1,-1);
      }
    }
    \node[font=\footnotesize] at (4,-9.4) {(a) Scaffolding encoder};
    \begin{scope}[xshift=5.6cm]
      \def\btype{{0,0,5,5,6,6,0,0,0,7,7,8,8,0,0}}
      \foreach \lab [count=\n from 0] in {$\mathbf{x}$,$\langle$BOT$\rangle$,$\mathbf{e}^{(1)}_{1}$,$\mathbf{e}^{(1)}_{2}$,$\mathbf{z}^{*(1)}_{1}$,$\mathbf{z}^{*(1)}_{2}$,$\langle$EOT$\rangle$,$\mathbf{c}$,$\langle$BOT$\rangle$,$\mathbf{e}^{(2)}_{1}$,$\mathbf{e}^{(2)}_{2}$,$\mathbf{z}^{*(2)}_{1}$,$\mathbf{z}^{*(2)}_{2}$,$\langle$EOT$\rangle$,$\mathbf{y}$}{
        \node[anchor=east] at (-0.1,-\n-0.5) {\lab};
        \node[anchor=west,rotate=90] at (\n+0.5,0.1) {\lab};
      }
      \foreach \i in {0,...,14}{
        \foreach \j in {0,...,14}{
          \pgfmathtruncatemacro{\ti}{\btype[\i]}
          \pgfmathtruncatemacro{\tj}{\btype[\j]}
          \pgfmathtruncatemacro{\isemb}{(\tj==5 || \tj==7) ? 1 : 0}
          \pgfmathtruncatemacro{\ok}{((\j<=\i) && (\isemb==0 || \ti==\tj)) || (\isemb==1 && \ti==\tj) ? 1 : 0}
          \ifnum\ok=1
            \ifnum\tj=5 \fill[purple!45] (\j,-\i) rectangle ++(1,-1);
            \else\ifnum\tj=7 \fill[purple!45] (\j,-\i) rectangle ++(1,-1);
            \else\ifnum\tj=6 \fill[oursred!70] (\j,-\i) rectangle ++(1,-1);
            \else\ifnum\tj=8 \fill[orange!70] (\j,-\i) rectangle ++(1,-1);
            \else \fill[bestblue!45] (\j,-\i) rectangle ++(1,-1);
            \fi\fi\fi\fi
          \else
            \fill[gray!10] (\j,-\i) rectangle ++(1,-1);
          \fi
          \draw[white,line width=0.3pt] (\j,-\i) rectangle ++(1,-1);
        }
      }
      \node[font=\footnotesize] at (7.5,-16.4) {(b) Predictor pass (Latent-Prediction Phase)};
      \node[font=\footnotesize] at (7.5,-17.6) {rows: queries; columns: keys};
    \end{scope}
  \begin{scope}[shift={(1.2,-19.2)},
      sw/.style={draw=none,minimum width=0.28cm,minimum height=0.28cm,inner sep=0pt},
      lt/.style={font=\footnotesize,inner sep=1pt,anchor=west}]
    \node[sw,fill=bestblue!45] (s1) at (0,0) {};  \node[lt,right=3pt of s1] (t1) {text / special tokens};
    \node[sw,fill=purple!45,right=14pt of t1] (s2) {}; \node[lt,right=3pt of s2] (t2) {learnable embeddings};
    \node[sw,fill=oursred!70,right=14pt of t2] (s3) {}; \node[lt,right=3pt of s3] (t3) {block 1};
    \node[sw,fill=orange!70,right=14pt of t3] (s4) {};  \node[lt,right=3pt of s4] (t4) {block 2};
    \node[sw,fill=gray!10,right=14pt of t4] (s5) {};    \node[lt,right=3pt of s5] (t5) {masked};
  \end{scope}
  \end{tikzpicture}}
  \caption{\textbf{Attention masks for processing all latent blocks of a trace in one forward pass} (two blocks, $K{=}2$, two tokens per helper image). \textbf{(a)}~Scaffolding encoder. \textbf{(b)}~Predictor pass in the Latent-Prediction Phase. $\mathbf{x}$: input; $\mathbf{c}$: text reasoning; $\mathbf{y}$: answer; $\mathbf{h}^{(b)}$: helper-image tokens; $\mathbf{e}^{(b)}$: learnable embeddings; $\mathbf{z}^{*(b)}$: target latent tokens. Colored cells are attended; gray cells are masked.}
  \label{fig:attn_mask}
\end{figure}

\paragraph{Attention masks.}\label{app:attn_mask}
A trace can contain several latent blocks, and computing them one after another would require one forward pass per block. We instead process all blocks of a trace in a single forward pass and use attention masks to preserve exactly the dependencies of the sequential computation (Figure~\ref{fig:attn_mask}). The masks follow two principles. \emph{Independence of targets:} in the scaffolding encoder, the helper images and learnable embeddings are packed as $[\mathbf{h}^{(1)},\mathbf{e}^{(1)},\mathbf{h}^{(2)},\mathbf{e}^{(2)},\ldots]$; each helper image attends only to itself, and each group of embeddings attends bidirectionally to itself and to the helper image right before it. Each target block $\mathbf{z}^{*(b)}$ therefore summarizes only its own helper image, exactly as if the images were encoded one at a time. \emph{No information beyond inference:} in the Latent-Prediction Phase, the model runs two forward passes. In the \emph{predictor pass}, it reads $[\mathbf{x},\langle\mathrm{BOT}\rangle,\mathbf{e}^{(1)},\mathbf{z}^{*(1)},\langle\mathrm{EOT}\rangle,\mathbf{c},\langle\mathrm{BOT}\rangle,\mathbf{e}^{(2)},\mathbf{z}^{*(2)},\langle\mathrm{EOT}\rangle,\mathbf{y}]$. Each group of embeddings attends bidirectionally within the group and sees the input and all preceding text and target blocks, but neither helper images nor the embeddings of earlier blocks, which do not exist at inference. No other token attends to the embeddings, so later blocks condition on the teacher-forced targets $\mathbf{z}^{*(b)}$, which the predicted blocks replace at inference (Appendix~\ref{app:teacher_forcing}). In the second, \emph{answer pass}, the same model reads $[\mathbf{x},\langle\mathrm{BOT}\rangle,\mathbf{z}^{*(1)},\langle\mathrm{EOT}\rangle,\mathbf{c},\ldots,\mathbf{y}]$ with its standard causal mask and generates the text from the latent blocks. In the Scaffolding Phase, the frozen VLM reads the same sequence with its standard causal mask.

\paragraph{Optimization.} Stage~1 (scaffolding encoder + latent generation) uses AdamW with learning rate $2{\times}10^{-5}$ for latent generation and $5{\times}10^{-5}$ for the scaffolding encoder, weight decay $0.01$, batch size $32$, cosine schedule with linear warmup over the first $5\%$ of steps, and BF16 mixed precision. We train for 3 epochs over the FrozenLake training set. Stage~2 (Scaffolding RL) uses GRPO with rollout group size $G{=}8$, clipping $\epsilon{=}0.2$, learning rate $1{\times}10^{-6}$, and KL penalty $\beta_{\mathrm{KL}}{=}0.001$ against the Stage~1 model; the mean head is zero-initialized and the variance head is initialized to $\sigma{=}0.05$. Latent generation loss weights are $\lambda_{\mathrm{latent}}{=}1.0$ and $\lambda_{\mathrm{task}}{=}0.5$ (default; see Appendix~\ref{sec:hyperparameter}).

\paragraph{Compute.} All experiments use NVIDIA A100 (80GB) GPUs. On FrozenLake, Stage~1 takes approximately $7$ GPU-hours per run and Stage~2 approximately $24$ GPU-hours per run, giving a total of ${\sim}31$ GPU-hours per configuration. Inference is performed on a single A100.

\paragraph{Evaluation.} We report exact-match accuracy: a generated path is correct only if it both reaches the goal and avoids every hole. The model outputs the full path as a sequence of directional moves (Up/\allowbreak Down/\allowbreak Left/\allowbreak Right); paths that pass through a hole are scored as zero reward, consistent with the VSP setup. All baselines are trained and evaluated by us under the identical data, backbone, and evaluation protocol.

{\subsection{Visual-Centric Reasoning Benchmarks}
\label{app:visual_setup}

\paragraph{Training data.} We train on Zebra-CoT~\citep{zebracot}: Stage~1 uses 162K interleaved traces and Stage~2 uses a disjoint 20K-trace split. Each trace already includes its intermediate helper images, which the scaffolding encoder consumes directly; we do not generate separate helper images. All nine benchmarks are evaluated on their full official test sets. No FrozenLake data are used in this experiment, and no Zebra-CoT data are used in the FrozenLake experiments.

\paragraph{Hyperparameters.} We reuse the same hyperparameters as for FrozenLake without further tuning: $K{=}4$ latent tokens, $\lambda_{\mathrm{latent}}{=}1.0$, $\lambda_{\mathrm{task}}{=}0.5$, AdamW with the same schedule, and the same Stage~2 GRPO setup. Training runs for 2 epochs over the Zebra-CoT training data. The full two-stage training takes approximately $200$ A100 GPU-hours.

\paragraph{Evaluation.} We follow each benchmark's standard protocol and report multiple-choice accuracy on all nine benchmarks.
}

\section{Helper Image Generation and Analysis}
\label{app:helper_code}

This appendix describes how the helper images used at training time are generated for each task family, lists the per-benchmark strategies, and analyzes the effect of helper image choice on FrozenLake.

\subsection{FrozenLake: Value-Function Heatmap}

\begin{verbatim}
import numpy as np

def value_iteration(grid, gamma=0.85, theta=1e-7):
    """Optimal state-value function for Frozen-Lake grid.
    Codes: 1=start, 2=goal, -1=hole, 0=safe."""
    n_rows, n_cols = grid.shape
    V = np.zeros_like(grid, dtype=float)
    actions = [(-1,0), (1,0), (0,-1), (0,1)]
    is_terminal = lambda r,c: grid[r,c] in (-1,2)
    while True:
        delta = 0.0
        for r in range(n_rows):
            for c in range(n_cols):
                if is_terminal(r,c): continue
                v_old = V[r,c]
                V[r,c] = max(
                    ((0. if grid[nr,nc]==-1 else
                      1. if grid[nr,nc]==2 else 0.)
                     + gamma * (0. if grid[nr,nc] in (-1,2)
                                else V[nr,nc]))
                    if 0<=nr<n_rows and 0<=nc<n_cols
                    else gamma * V[r,c]
                    for dr,dc in actions
                    for nr,nc in [(r+dr, c+dc)])
                delta = max(delta, abs(v_old - V[r,c]))
        if delta < theta: break
    V[grid == -1] = 0.0; V[grid == 2] = 1.0
    return V
\end{verbatim}

{\subsection{Visual-Centric Reasoning Benchmarks}
\label{app:vision_helpers}

For visual-centric reasoning benchmarks, every Zebra-CoT training example already contains one or more intermediate helper images paired with its query~\citep{zebracot}. The scaffolding encoder consumes these provided helper images directly during Stage~1 training; we do not generate a separate helper image for these tasks.
}

\section{Additional Experimental Results}
\label{app:additional_results}

\postponedablations

\subsection{Additional Ablations}
\label{app:more_ablations}

\paragraph{Parameter-matched control.} The scaffolding encoder tunes a copy of the vision encoder, so its gain could come from added trainable parameters rather than from a better latent target. To test this on FrozenLake, we fine-tune the same vision encoder in plain SFT without any latent target. Tuning the encoder alone raises SFT from $52.0\%$ to $56.5\%$ (90/72/44/20 on L8/16/24/32), whereas learning the latent target with the same tuned encoder reaches $72.0\%$. Added capacity therefore explains only a small part of the gain. Consistently, a frozen encoder cannot exploit a richer helper image (Table~\ref{tab:helper_choice}). The scaffolding encoder, a ${\sim}677$M-parameter copy of the vision tower plus a ${\sim}80$M-parameter attention-pooling module (${\sim}0.76$B in total, ${\sim}9.1\%$ of the 8.29B backbone), is used only during training and removed at inference.

\paragraph{Stage~1 and Stage~2 are complementary.} To test whether Stage~2 can compensate for a weak latent target, we apply the three Stage~2 recipes to two Stage~1 checkpoints: ours ($72.0\%$) and a prior-work checkpoint with a frozen target (mean pooling) ($57.8\%$) (Figure~\ref{fig:rl_full} and Table~\ref{tab:rl_ablation}). On the weak checkpoint, Scaffolding RL still gives the largest gain ($+5.2$ points, versus $+1.2$ points for VLPO and $+0.6$ points for text-only GRPO), even larger than its $+3.0$ points on our checkpoint because there is more headroom. Yet it reaches only $63.0\%$, below even the weakest recipe on top of the scaffolding encoder ($72.6\%$). Stage~2 therefore refines the latent target that Stage~1 provides but cannot replace it, and the two stages are complementary.

\paragraph{Ablation on visual-centric benchmarks.} To test whether both components help beyond FrozenLake, we repeat the two key FrozenLake ablations on the nine visual-centric benchmarks. All variants use Qwen2.5-VL-7B, the Zebra-CoT data of Section~\ref{sec:perception}, $K{=}4$, and the hyperparameters of Appendix~\ref{app:visual_setup}. In Stage~1, mirroring the SFT visual target ablation (Figure~\ref{fig:ablation}), both variants are trained on the same 162K SFT traces with the same latent predictor and differ only in the latent target: a frozen target from mean-pooled vision-encoder features, as in prior work, or targets from our scaffolding encoder. In Stage~2, mirroring Table~\ref{tab:rl_ablation}, both variants start from our Stage~1 checkpoint ($70.9\%$ average; Table~\ref{tab:perception}), use the same 20K RL traces and rollout budget, and differ only in the RL recipe. The scaffolding encoder raises the Stage~1 average from $66.3\%$ to $70.9\%$, and Scaffolding RL raises the Stage~2 average from $72.7\%$ (VLPO) to $73.9\%$, improving every benchmark in both cases (Table~\ref{tab:zebra_ablation}). Both components therefore transfer beyond spatial planning.

\begin{table}[H]
\centering
\caption{\textbf{Ablation on the visual-centric reasoning benchmarks} (accuracy \%). All rows use the same backbone and Zebra-CoT training data. Stage~1 rows differ only in the latent target; Stage~2 rows start from the same Scaffolding Encoder checkpoint and differ only in the RL recipe.}
\label{tab:zebra_ablation}
\setlength{\tabcolsep}{3.6pt}
\renewcommand{\arraystretch}{1.1}
\resizebox{\linewidth}{!}{%
\begin{tabular}{ll|ccccccccc|c}
\toprule
\rowcolor{headerbg}
& \textbf{Variant} & \textbf{V$^\star$} & \textbf{BLINK} & \textbf{MMVP} & \textbf{MMStar} & \textbf{CVB.} & \textbf{HRB-4K} & \textbf{HRB-8K} & \textbf{MME} & \textbf{Jig.} & \textbf{Avg.} \\
\midrule
\multirow{2}{*}{Stage~1} & Frozen target & 82.4 & 57.5 & 71.3 & 67.8 & 79.0 & 72.0 & 68.3 & 45.2 & 52.8 & 66.3 \\
 & \cellcolor{oursbg}\textbf{Scaffolding encoder} & \cellcolor{oursbg}\textbf{87.4} & \cellcolor{oursbg}\textbf{64.8} & \cellcolor{oursbg}\textbf{73.3} & \cellcolor{oursbg}\textbf{72.1} & \cellcolor{oursbg}\textbf{86.4} & \cellcolor{oursbg}\textbf{73.6} & \cellcolor{oursbg}\textbf{71.1} & \cellcolor{oursbg}\textbf{53.4} & \cellcolor{oursbg}\textbf{56.0} & \cellcolor{oursbg}\textbf{70.9} \\
\midrule
\multirow{2}{*}{Stage~2} & VLPO & 89.3 & 66.2 & 75.3 & 72.9 & 87.6 & 75.9 & 72.9 & 55.7 & 58.2 & 72.7 \\
 & \cellcolor{oursbg}\textbf{Scaffolding RL} & \cellcolor{oursbg}\textbf{90.6} & \cellcolor{oursbg}\textbf{67.2} & \cellcolor{oursbg}\textbf{76.7} & \cellcolor{oursbg}\textbf{73.5} & \cellcolor{oursbg}\textbf{88.4} & \cellcolor{oursbg}\textbf{77.5} & \cellcolor{oursbg}\textbf{74.1} & \cellcolor{oursbg}\textbf{57.3} & \cellcolor{oursbg}\textbf{59.7} & \cellcolor{oursbg}\textbf{73.9} \\
\bottomrule
\end{tabular}}
\end{table}

\subsection{Teacher-Forced vs.\ Self-Predicted Latents}
\label{app:teacher_forcing}
In the Latent-Prediction Phase of Stage~1, the model runs two forward passes per trace: a \emph{predictor pass} that generates all latent blocks, and an \emph{answer pass} of the same model that generates the text from them. We train both with teacher forcing. This is convenient, but it raises an obvious question. During training, the predictor pass computes each latent block from a context that contains the preceding \emph{target} blocks $\mathbf{z}^*$, and the answer pass conditions on these targets as well; at inference, both only see the blocks the model predicted itself. If the two differ noticeably, the model may learn to rely on context it never gets at test time. We test this directly. As a reference, we train a self-predicted variant in which later tokens condition on the preceding \emph{predicted} blocks $\hat{\mathbf{z}}$ during training as well, with a stop-gradient $\mathrm{sg}(\hat{\mathbf{z}})$ so that no gradient flows back through this context. We run the comparison in the interleaved visual-centric setting, where each latent block builds on the ones before it, so any mismatch would accumulate across blocks. Everything else (the Zebra-CoT data, the backbone, and the hyperparameters) stays the same, and we evaluate the resulting Stage~1 models on four benchmarks. In terms of accuracy, the two variants end up almost identical: on every benchmark the difference is at most $0.4$ points (Table~\ref{tab:teacher_forcing}). The cost, however, is quite different. The self-predicted variant has to finish each latent block before it can predict the next one, so it needs $B$ sequential predictor passes instead of one, that is, $B{+}1$ forward passes per trace in total instead of $2$, and in practice it trains about $3.2\times$ slower. In short, teacher forcing introduces no noticeable mismatch here, since the matching loss already keeps the predicted blocks close to their targets, and it is much cheaper to train. We therefore use it in all experiments.

\begin{table}[H]
\centering
\caption{\textbf{Teacher-forced vs.\ self-predicted latent context} (Stage~1, accuracy \%). \emph{Forward passes}: predictor passes $+$ answer pass per training trace with $B$ latent blocks. \emph{Training time}: Stage~1 training time relative to teacher forcing.}
\label{tab:teacher_forcing}
\setlength{\tabcolsep}{4.5pt}
\renewcommand{\arraystretch}{1.1}
\resizebox{\linewidth}{!}{%
\begin{tabular}{l|cccc|c|cc}
\toprule
\rowcolor{headerbg}
\textbf{Preceding latent context} & \textbf{V$^\star$} & \textbf{MMVP} & \textbf{CVB.} & \textbf{HRB-4K} & \textbf{Mean} & \textbf{Forward passes} & \textbf{Training time} \\
\midrule
\rowcolor{oursbg}
Target $\mathbf{z}^*$ (teacher forcing, ours) & 87.4 & 73.3 & 86.4 & 73.6 & 80.2 & $1{+}1$ & $1.0\times$ \\
Predicted $\mathrm{sg}(\hat{\mathbf{z}})$ (self-predicted) & 87.1 & 72.9 & 86.2 & 73.3 & 79.9 & $B{+}1$ & $3.2\times$ \\
\bottomrule
\end{tabular}}
\end{table}

\subsection{RL Exploration Analysis}
\label{app:rl_diag}
Table~\ref{tab:rl_ablation} and Figure~\ref{fig:rl_full} show that Scaffolding RL gains more, and more stably, than VLPO and text-only GRPO. Here we examine how the three recipes explore during training to identify where this advantage comes from. All recipes start from the same Stage~1 checkpoint on FrozenLake ($72.0\%$) with the same rollout budget, and we track three metrics (Table~\ref{tab:rl_diag}). The \emph{reward standard deviation within a rollout group} measures how different the outcomes of the rollouts for the same prompt are; GRPO's group-relative advantage vanishes when all rollouts receive the same reward, so a larger spread gives a more informative update. \emph{Pass@$k$} is the percentage of training prompts for which at least one of $k$ rollouts is correct, and measures whether exploration reaches correct solutions. The \emph{latent Gaussian scale} $\sigma$ indicates how far latent samples move from the predicted latent. Text-only GRPO explores only through text sampling and has the lowest reward spread ($0.15$) and pass@$16$ ($80$). VLPO also samples only text, since it regularizes the latent block with a fixed scale instead of sampling it; its spread ($0.32$) is higher than that of text-only GRPO, because updating the latent block also shifts the text rollouts that condition on it, but it has no latent exploration to adapt. Scaffolding RL samples latent adjustments with a learned, input-dependent scale ($0.04$--$0.15$), which gives the largest reward spread ($0.42$) and the highest pass@$k$ for every $k$. The advantage of Scaffolding RL therefore comes from latent exploration: sampling alternative latent reasoning paths yields more diverse and more often successful rollouts, which gives GRPO a stronger learning signal, and the learned scale keeps this exploration adaptive.

\begin{table}[H]
\centering
\caption{\textbf{RL exploration metrics} on FrozenLake, all from the same Stage~1 checkpoint ($72.0\%$). $^\ast$VLPO does not sample latents; $0.05$ is the fixed scale $\sigma_0$ of its Gaussian regularizer (Section~\ref{sec:prelim}).}
\label{tab:rl_diag}
\setlength{\tabcolsep}{4.5pt}
\renewcommand{\arraystretch}{1.05}
\begin{tabular}{l|ccc}
\toprule
\rowcolor{headerbg}
\textbf{Metric} & \textbf{Text-only GRPO} & \textbf{VLPO} & \textbf{Scaffolding RL} \\
\midrule
Reward s.d.\ within a rollout group & 0.15 & 0.32 & \textbf{0.42} \\
Pass@4 / @8 / @16 (\%) & 74 / 76 / 80 & 75 / 79 / 82 & \textbf{79 / 82 / 85} \\
Latent Gaussian scale $\sigma$ & N/A & 0.05$^\ast$ (fixed) & 0.04--0.15 (learned) \\
\bottomrule
\end{tabular}
\end{table}

\subsection{Latent Generation Loss Weights}
\label{sec:hyperparameter}

We sweep the matching weight $\lambda_{\mathrm{latent}}$ and the task weight $\lambda_{\mathrm{task}}$ of the Latent-Prediction Phase objective on FrozenLake (Levels~8, 16, 24, and 32; Table~\ref{tab:hyper}); $\lambda_{\mathrm{latent}}$ is kept positive, since the matching term is what trains the latent prediction. Without the task term ($\lambda_{\mathrm{task}}{=}0$, $64.8\%$), the model only learns to reproduce $\mathbf{z}^*$ and not to answer from it, which is the weakest setting. Adding the task term improves every level, and the default $\lambda_{\mathrm{latent}}{=}1.0$, $\lambda_{\mathrm{task}}{=}0.5$ gives the best average accuracy ($72.0\%$): the matching term keeps the predicted latent close to its target, and the task term trains the same model to answer from the latent blocks.

\begin{table}[H]
\centering
\captionsetup{font=normalsize}
\caption{\textbf{Latent generation loss weights} on FrozenLake (accuracy \%; Avg.\ over the four levels shown). $\lambda_{\text{latent}}$ and $\lambda_{\text{task}}$ weight the latent matching term and the task loss term, respectively.}
\label{tab:hyper}
\normalsize
\setlength{\tabcolsep}{5pt}
\renewcommand{\arraystretch}{1.05}
\begin{tabular}{cc|cccc|c}
\toprule
\rowcolor{headerbg}
$\lambda_{\text{latent}}$ & $\lambda_{\text{task}}$ & \textbf{L8} & \textbf{L16} & \textbf{L24} & \textbf{L32} & \textbf{Avg.} \\
\midrule
1.0 & 0.0 & 88.0 & 72.0 & 54.0 & 45.0 & 64.8 \\
\rowcolor{oursbg}
\textbf{1.0} & \textbf{0.5} & \textbf{94.0} & \textbf{79.0} & \textbf{63.0} & \textbf{52.0} & \textbf{72.0} \\
1.0 & 1.0 & 92.0 & 77.0 & 60.0 & 51.0 & 70.0 \\
0.5 & 1.0 & 93.0 & 78.0 & 62.0 & 52.0 & 71.3 \\
\bottomrule
\end{tabular}
\end{table}

\subsection{Effect of Model Scale and LoRA Fine-tuning}
\label{app:scale_lora}

To check that the framework's gains are not specific to the 7B backbone or to full fine-tuning, we repeat the FrozenLake experiment under two variants: a smaller Qwen2.5-VL-3B backbone, and a parameter-efficient LoRA setup on the same Qwen2.5-VL-7B backbone (rank $r{=}32$, $\alpha{=}64$, dropout $0.05$, applied to all attention and MLP projection matrices). Both runs use the same data, helper images, hyperparameters, and evaluation protocol as the main run; only the backbone size or trainability scope changes. Table~\ref{tab:scale_lora} reports the joint result. At 3B scale, Scaffolding Encoder+RL improves over the 3B SFT baseline by $+17.2$ points on average, and the gap is largest on the harder levels ($+22$ points at Level~24 and $+18$ points at Level~32); its $+5.7$-point improvement over the strongest prior latent baseline is preserved. The LoRA setup also improves by $+17.0$ points over LoRA SFT, although its absolute average remains below 3B full fine-tuning ($58.8\%$ vs.\ $64.0\%$). Full fine-tuning at 7B remains the strongest configuration.

\begin{table}[H]
  \centering
  \caption{\textbf{Effect of model scale and LoRA fine-tuning on FrozenLake} (accuracy \%). Left: Qwen2.5-VL-3B with full fine-tuning. Right: Qwen2.5-VL-7B with LoRA $r{=}32$. }
  \label{tab:scale_lora}
  \setlength{\tabcolsep}{3.5pt}
  \renewcommand{\arraystretch}{1.10}
  \resizebox{\linewidth}{!}{%
  \begin{tabular}{l|cccc|c|cccc|c}
  \toprule
  \rowcolor{headerbg}
  & \multicolumn{5}{c|}{\textbf{Qwen2.5-VL-3B (full FT)}} & \multicolumn{5}{c}{\textbf{Qwen2.5-VL-7B (LoRA $r{=}32$)}} \\
  \textbf{Method} & \textbf{Lv.\ 8} & \textbf{Lv.\ 16} & \textbf{Lv.\ 24} & \textbf{Lv.\ 32} & \textbf{Avg.} & \textbf{Lv.\ 8} & \textbf{Lv.\ 16} & \textbf{Lv.\ 24} & \textbf{Lv.\ 32} & \textbf{Avg.} \\
  \midrule
  SFT                            & 75.0 & 60.0 & 38.0 & 14.0 & 46.8 & 70.0 & 52.0 & 32.0 & 13.0 & 41.8 \\
  SFT+GRPO                       & 77.0 & 62.0 & 40.0 & 16.0 & 48.8 & 71.0 & 54.0 & 34.0 & 14.0 & 43.3 \\
  LVR~\citep{li2025lvr}           & 78.0 & 63.0 & 41.0 & 18.0 & 50.0 & 73.0 & 56.0 & 36.0 & 16.0 & 45.3 \\
  Mirage~\citep{yang2025b}        & 79.0 & 65.0 & 43.0 & 20.0 & 51.8 & 74.0 & 58.0 & 38.0 & 18.0 & 47.0 \\
  VaLR~\citep{jeon2026valr}       & 83.0 & 71.0 & 51.0 & 28.0 & 58.3 & 80.0 & 65.0 & 44.0 & 24.0 & 53.3 \\
  \rowcolor{oursbg}
  \textbf{Scaffolding Encoder}
  & 86.0
  & 73.0
  & 56.0
  & 30.0
  & 61.3
  & \textbf{84.0}
  & 64.0
  & 48.0
  & 32.0
  & 57.0 \\
  
  \rowcolor{oursbest}
  \textbf{Scaffolding Encoder+RL}
  & \textbf{88.0}
  & \textbf{76.0}
  & \textbf{60.0}
  & \textbf{32.0}
  & \textbf{64.0}
  & 82.0
  & \textbf{67.0}
  & \textbf{51.0}
  & \textbf{35.0}
  & \textbf{58.8} \\
  \bottomrule
  \end{tabular}}
\end{table}

\subsection{Inference Efficiency}
\label{app:efficiency}
We measure whether latent reasoning keeps inference close to the cost of the base VLM, both when a trace has a single latent block (FrozenLake) and when latent blocks are interleaved with text (the visual-centric benchmarks). We report single-query latency on one A100 in BF16, averaged over 100 queries (Table~\ref{tab:efficiency}). At inference, the scaffolding encoder is removed; each latent block adds only $K{=}4$ positions computed in one forward pass, and the two Gaussian heads are small MLPs. Our method adds $+4.6\%$ latency on FrozenLake and $+5.9\%$ on the visual-centric benchmarks, within the range of other latent reasoning methods ($+2.3\%$ to $+8.4\%$). Tool-calling and image-generation methods add $+76\%$ to $+277\%$ because they process or generate explicit images at each reasoning step. Our method thus stays near base-VLM cost in both single-block and interleaved settings.

\begin{table}[ht]
\centering
\caption{\textbf{Single-query latency} (single A100, BF16, averaged over 100 queries). $K$: latent tokens per block (--- for methods that process or generate images). --: not measured.}
\label{tab:efficiency}
\setlength{\tabcolsep}{5pt}
\renewcommand{\arraystretch}{1.1}
\resizebox{\linewidth}{!}{%
\begin{tabular}{l|c|cc|cc}
\toprule
\rowcolor{headerbg}
& & \multicolumn{2}{c|}{\textbf{FrozenLake (single block)}} & \multicolumn{2}{c}{\textbf{Visual-centric (interleaved)}} \\
\rowcolor{headerbg}
\textbf{Method} & $\boldsymbol{K}$ & \textbf{Latency (ms)} & \textbf{Overhead} & \textbf{Latency (ms)} & \textbf{Overhead} \\
\midrule
Qwen2.5-VL-7B (base) & --- & 260 & --- & 740 & --- \\
\midrule
\rowcolor{basebg}
\multicolumn{6}{c}{\textit{\small Latent reasoning methods}} \\
Mirage~\citep{yang2025b} & 4 & 266 & $+2.3\%$ & -- & -- \\
LVR~\citep{li2025lvr} & 4 & 267 & $+2.7\%$ & 794 & $+7.3\%$ \\
VaLR~\citep{jeon2026valr} & 4 & 268 & $+3.1\%$ & 802 & $+8.4\%$ \\
SkiLa~\citep{tong2025skila} & 4 & 269 & $+3.5\%$ & -- & -- \\
CoVT~\citep{qin2025covt} & 8 & 274 & $+5.4\%$ & -- & -- \\
Monet~\citep{wang2025monet} & 8 & 276 & $+6.2\%$ & -- & -- \\
\rowcolor{oursbg}
\textbf{Scaffolding Encoder+RL (Ours)} & 4 & 272 & $+4.6\%$ & 784 & $+5.9\%$ \\
\midrule
\rowcolor{basebg}
\multicolumn{6}{c}{\textit{\small Tool-calling \& image-generation methods}} \\
DeepEyes~\citep{zheng2025deepeyes} & --- & ${\sim}540$ & $+108\%$ & 1358 & $+83.5\%$ \\
Thyme~\citep{zhang2025thyme} & --- & ${\sim}680$ & $+162\%$ & 1305 & $+76.4\%$ \\
VPRL~\citep{xu2025} & --- & ${\sim}820$ & $+215\%$ & -- & -- \\
DiffThinker~\citep{xu2025diffthinker} & --- & ${\sim}910$ & $+250\%$ & -- & -- \\
MVoT~\citep{li2025mvot} & --- & ${\sim}980$ & $+277\%$ & -- & -- \\
\bottomrule
\end{tabular}}
\end{table}

\end{document}